\PassOptionsToPackage{dvipsnames}{xcolor}
\documentclass[singlecolumn]{ModernAcademic-EN}

\setthemecolor{224488}
\settitlepreskip{1ex}
\settitlelogogap{0.8em}

\usepackage{array}

\usepackage{mathtools}

\usepackage[abbreviations]{glossaries-extra}
\glsdisablehyper
\setabbreviationstyle{long-short}
\newabbreviation{SOD}{SOD}{Salient Object Detection}
\newabbreviation{COD}{COD}{Camouflaged Object Detection}
\newabbreviation{TOS}{TOS}{Transparent Object Segmentation}
\newabbreviation{GD}{GD}{Glass Detection}
\newabbreviation{MD}{MD}{Mirror Detection}
\newabbreviation{SD}{SD}{Shadow Detection}
\newabbreviation{DIS}{DIS}{Dichotomous Image Segmentation}
\newabbreviation{MLS}{MLS}{Medical Lesion Segmentation}
\newabbreviation{IRSTD}{IRSTD}{Infrared Small Target Detection}
\newabbreviation{GT}{GT}{Ground Truth}
\newabbreviation{MAE}{MAE}{Mean Absolute Error}
\newabbreviation{OA}{OA}{Overall Accuracy}
\newabbreviation{IoU}{IoU}{Intersection-over-Union}
\newabbreviation{hIoU}{hIoU}{Hierarchical IoU}
\newabbreviation{SIMAE}{SI-MAE}{Size-Invariant Mean Absolute Error}
\newabbreviation{Fbeta}{\ensuremath{F_{\beta}}}{F-measure}
\newabbreviation{wFbeta}{\ensuremath{F_{\beta}^{\omega}}}{Weighted F-measure}
\newabbreviation{Sm}{\ensuremath{S_m}}{Structure Measure}
\newabbreviation{Em}{\ensuremath{E_m}}{Enhanced Alignment Measure}
\newabbreviation{MSIoU}{MSIoU}{Multi-Scale IoU}
\newabbreviation{ETMSIoU}{ETMSIoU}{Equal-Target Multiscale Soft \glsentryshort{IoU}}
\newabbreviation{HCE}{HCE}{Human Correction Effort}
\newabbreviation{Cm}{\ensuremath{C_{\beta}}}{Context Measure}
\newabbreviation{wCm}{\ensuremath{C_{\beta}^{\omega}}}{Weighted Context Measure}
\newabbreviation{ASD}{ASD}{Average Surface Distance}
\newabbreviation{ASSD}{ASSD}{Average Symmetric Surface Distance}
\newabbreviation{HD}{HD}{Hausdorff Distance}
\newabbreviation{HD95}{HD95}{95th-Percentile Hausdorff Distance}
\newabbreviation{SegAE}{SegAE}{Segmentation Assessment Engine}
\newabbreviation{FT}{FT}{Fractional Threshold}
\newabbreviation{SSIM}{SSIM}{Structural Similarity Index}
\newabbreviation{TP}{TP}{True Positive}
\newabbreviation{FP}{FP}{False Positive}
\newabbreviation{TN}{TN}{True Negative}
\newabbreviation{FN}{FN}{False Negative}
\newabbreviation{TPR}{TPR}{True Positive Rate}
\newabbreviation{TNR}{TNR}{True Negative Rate}
\newabbreviation{FPR}{FPR}{False Positive Rate}
\newabbreviation{FNR}{FNR}{False Negative Rate}
\newabbreviation{BA}{BA}{Balanced Accuracy}
\newabbreviation{BER}{BER}{Balanced Error Rate}
\newabbreviation{AUC}{AUC}{Area Under the Curve}
\newabbreviation{AP}{AP}{Average Precision}
\newabbreviation{ROC}{ROC}{Receiver Operating Characteristic}
\glsunset{Fbeta}

\newcommand{\parhead}[1]{\noindent\textbf{\textit{#1}}}

\usepackage{adjustbox}

\usepackage{titletoc}

\newcommand{\setrefname}[3]{\crefname{#1}{#2}{#3}\Crefname{#1}{#2}{#3}}
\setrefname{figure}{Fig.}{Figs.}
\setrefname{subfigure}{Fig.}{Figs.}
\setrefname{table}{Tab.}{Tabs.}
\setrefname{equation}{Eq.}{Eqs.}
\setrefname{section}{Sec.}{Secs.}
\setrefname{subsection}{Sec.}{Secs.}
\setrefname{subsubsection}{Sec.}{Secs.}

\newtcolorbox{scopebox}{
  breakable,
  colback=black!4, colframe=black!50, boxrule=0.4pt, arc=0pt, outer arc=0pt,
  boxsep=0pt, left=5pt, right=5pt, top=4pt, bottom=4pt,
  before skip=8pt plus 2pt, after skip=8pt plus 2pt}

\definecolor{entryname}{HTML}{9A4A2C}
\newcommand{\ptype}[1]{Type \textcolor{entryname}{#1}}
\newcommand{\trow}[1]{Row \textcolor{entryname}{#1}}
\newcommand{\tcol}[1]{Column \textcolor{entryname}{#1}}
\newcommand{\tgroup}[1]{Group \textcolor{entryname}{#1}}
\newcommand{\tname}[1]{\textcolor{entryname}{#1}}
\newcommand{\diaghead}[1]{\makebox[0pt][l]{\rotatebox{45}{#1}}}
\newcommand{\tsd}[1]{{\scriptsize\textcolor{ForestGreen}{$\pm#1$}}}

\usepackage{xspace}
\DeclareRobustCommand{\ie}{\textit{i.e.}\@\xspace}

\DeclareRobustCommand{\vs}{\textit{vs.}\@\xspace}

\title{From Formulas to Protocols: A Compositional View of Binary Target Segmentation Metrics}
\shorttitle{From Formulas to Protocols}
\author{Youwei Pang, Xiaoqi Zhao}
\shortauthor{Youwei \& Xiaoqi}
\affiliation{Nanyang Technological University}
\email{lartpang@gmail.com}

\abstract{%
  Evaluation metrics are central to binary target segmentation because they determine how progress is measured, compared, and interpreted.
  In this paper, target denotes the task-defined positive region to be segmented rather than a generic foreground object.
  It may be salient, camouflaged, transparent, glass-like, mirror-like, shadow-like, lesion-like, or defined by other application-specific semantics.
  We treat metrics as compositions of modular design choices rather than isolated formulas.
  The proposed framework decomposes each metric into the five stages of prediction representation, target extraction, target matching, score computation, and metric reporting.
  Analyzing representative metrics by stage keeps their assumptions visible and shows how newer metrics address specific limits in earlier protocols.
  Controlled examples show how protocol choices affect evaluation outcomes and how existing mechanisms can be recombined for specific evaluation needs.
  We then discuss the resulting design space and its implications for task-aware evaluation protocols.
  Reference code is available at \url{https://github.com/lartpang/PySODMetrics}.
}

\begin{document}

\maketitle

\section{Introduction}

Binary target segmentation is a common output format in many vision tasks, including \gls{SOD}~\cite{CMMSODSurvey,RGBDSODSurveyZhou,DUTS,DUT-OMRON,MINet,PoolNet,GateNet,HDFNet,DANet,SOD-S3OD},
\gls{COD}~\cite{ConcealedObjectDetection,CAMO,COD10K,COD-PFNet,ZoomNet-CVPR2022,ZoomNeXt},
\gls{TOS}~\cite{TransparentObjectSegmentation},
\gls{GD}~\cite{GlassDetection},
\gls{MD}~\cite{MirrorDetection},
\gls{SD}~\cite{SurveyShadowDetection,CDConceptSeg-Spider},
\gls{DIS}~\cite{HumanCorrectionEffortMeasure,DIS-BiRefNet,BASNet,DIS-MVANet},
\gls{MLS}~\cite{PraNet,MSNet,M2SNet},
and others~\cite{Survey-RSCD,IRSTDSurvey}.
We use \emph{target} as an umbrella term for the task-defined positive region to be segmented.
Such a target is not necessarily a perceptually salient foreground object.
It may be hidden, transparent, reflective, cast by illumination, medically abnormal, changed between observations, or otherwise defined by task semantics.
Although such tasks have different visual goals, they share the same basic evaluation setting in which a predicted continuous target probability map is compared with a binary \gls{GT} mask.
The metric used here is not a neutral technical choice.
It determines which errors are visible, which model behaviors are rewarded, and how reliably methods can be compared across datasets with different target sizes, boundary complexity, and target-background contrast.

\begin{figure}[t!]
  \centering
  \includegraphics[width=\linewidth]{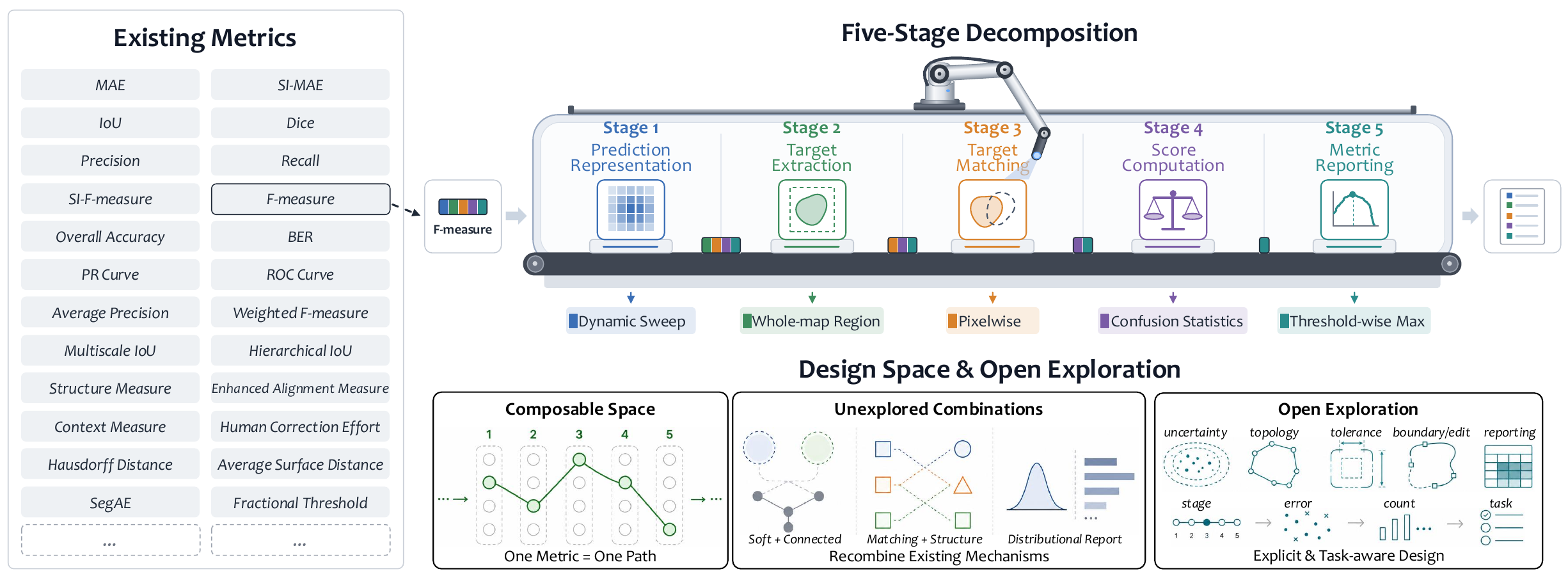}
  \caption{From isolated metrics to a five-stage framework.
    Existing segmentation metrics are commonly introduced as independent names or formulas, hiding the protocol choices that make them comparable or different.
    We decouple them into five stages, which localizes shared operations and metric-specific assumptions, reveals underexplored combinations, and gives new task-aware metrics a stage-wise rationale.
    Metric innovation then shifts from inventing isolated formulas to working in an explicit design space.}
  \label{fig:core_idea}
\end{figure}

Early target-map evaluation relied heavily on pixelwise error and overlap-based measures, often under the historical terminology of foreground-map evaluation.
These metrics remain useful because they are simple, interpretable, and efficient.
However, pixelwise and overlap scores usually treat pixels as independent counting units.
As a result, they can underrepresent target structure, topology, boundary quality, contextual plausibility, target-size fairness, and human editing cost.
These limitations have led to broader metric families, including weighted~\cite{wFmeasure,ContextMeasure}, structure-aware~\cite{Smeasure}, alignment-based~\cite{Emeasure}, multiscale~\cite{MSIoU}, effort-oriented~\cite{HumanCorrectionEffortMeasure}, size-invariant~\cite{SizeInvarianceVariants}, and context-aware metrics~\cite{ContextMeasure}.
This broader set covers more failure modes, but it also makes evaluation harder to organize.
Each metric is often introduced with its own notation, calculation details, and task motivation, so differences in presentation can obscure the shared operations underneath.
Metrics with similar final formulas may differ in thresholding, target representation, matching, aggregation, or reporting.
Metrics with different formulas may share most of their upstream evaluation path.
Thus, metric comparison requires more than listing equations.
It requires exposing the computational path that turns a prediction and a \gls{GT} mask into a reported number.

We analyze representative binary target segmentation metrics as modular compositions.
Every metric must decide how the prediction is represented, what target is extracted, how prediction and \gls{GT} are put into correspondence, how the score is computed, and how the final result is reported.
We decompose metric computation into five stages covering prediction representation, target extraction, target matching, score computation, and metric reporting.
The decomposition separates shared evaluation operations from metric-specific design choices, as illustrated in \cref{fig:core_idea}.
We make four contributions.
\textbf{First}, we propose a modular five-stage framework that decomposes binary target segmentation metrics into their constituent computational choices.
\textbf{Second}, we instantiate the framework on representative metrics, showing that their apparent diversity often reflects different compositions of a shared evaluation pipeline.
\textbf{Third}, we use the stage mapping to make metric assumptions explicit, including how thresholding, target granularity, matching, scoring, and reporting choices affect what a metric measures.
\textbf{Fourth}, we demonstrate metric construction through a worked combination of stage mechanisms and summarize the reporting conditions needed for reproducible comparison.

\begin{scopebox}
\textbf{Scope of This Work.}
We contribute a conceptual framework that treats each metric as a path through a compositional evaluation pipeline.
Our focus is on establishing a coherent and reusable abstraction for understanding existing metrics and guiding future metric design, rather than introducing a new model or dataset evaluated through numerical gains.
An illustrative metric construction shows how explicit evaluation requirements can be implemented by combining stage mechanisms.
We use this abstraction to expose protocol assumptions, explain relationships among metrics, and identify how their constituent mechanisms can be recombined for different evaluation needs.
The controlled examples make these relationships and design choices concrete.
\end{scopebox}

\begin{figure*}[t]
  \centering
  \includegraphics[width=\linewidth]{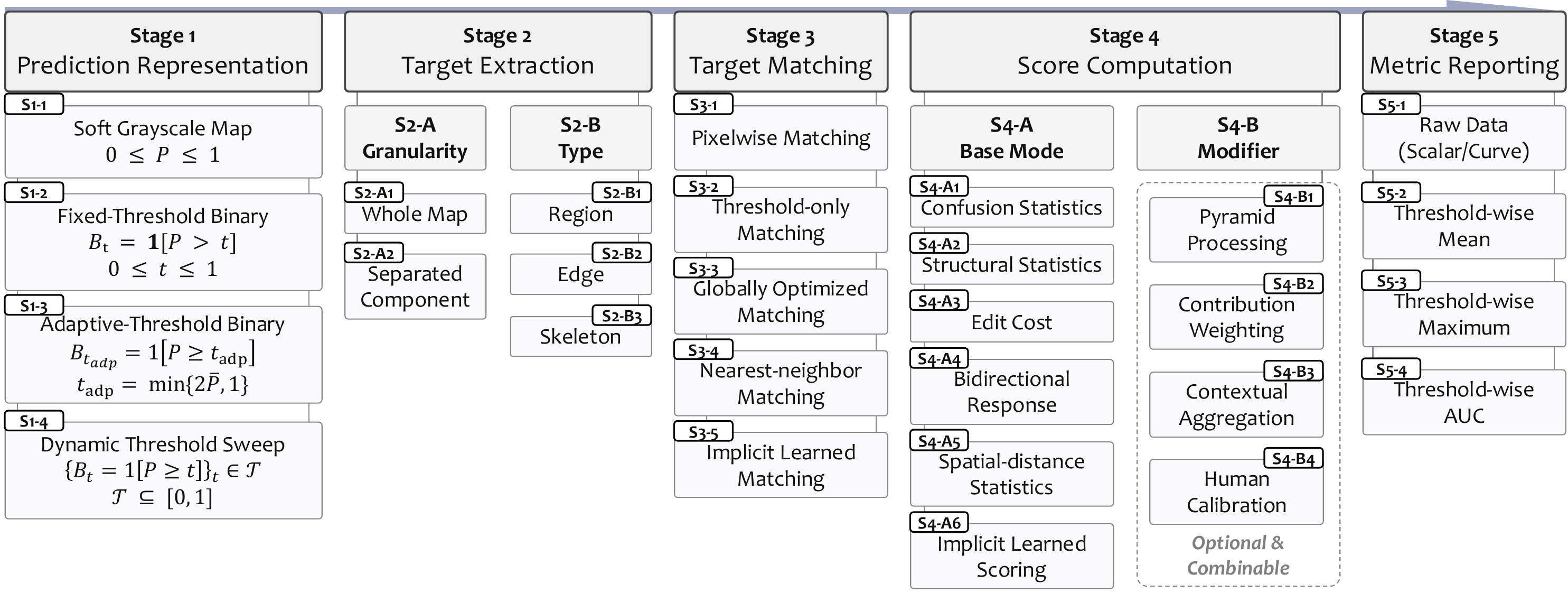}
  \caption{A decoupled five-stage view of binary target segmentation metrics.
  S2 specifies target granularity and type.
  S4 specifies a base scoring mode and optional modifiers.
  Pyramid processing includes scale combination.}
  \label{fig:metric_pipeline}
\end{figure*}

\section{A Five-Stage Framework for Metric Design}
\label{sec:five_stage_framework}

\subsection{Metrics as Flexible Compositions}
\label{subsec:framework_overview}

A score reported by existing binary target segmentation metrics usually combines design choices from several stages.

\parhead{Stage Boundaries.}
We distinguish stages by what an operation changes, \ie, the prediction representation (S1), evaluated entities (S2), correspondence relation (S3), quality score (S4), or final report (S5), as reflected in~\cref{fig:metric_pipeline}.
Thresholding belongs to S1 because it changes the prediction representation and can consequently create, split, or merge the connected targets extracted in S2.
Entity extraction, matching, and scoring remain distinct choices because they determine what is evaluated, which entities are compared, and how their agreement is measured, even when an implementation combines these operations.
The stage view makes metric design modular.
A metric is one path through five option sets rather than a single indivisible equation.

\subsection{S1: Prediction Representation}
\label{subsec:stage_prediction_treatment}

The first stage decides how the prediction map enters evaluation.
This decision directly changes the evaluation question.
Keeping the soft map tests the quality of confidence values, using a fixed threshold tests one operating point, and sweeping thresholds tests how stable the prediction ranking is across operating points.
Unless otherwise stated, the discussion assumes a binary mask $G$ and a target-channel prediction with $0\le P(x)\le1$ at each pixel $x$, such as a sigmoid output or the target-class softmax probability of a two-logit output.
A prediction stored as an 8-bit map $Q$ enters evaluation as $P=Q/255$.
Soft protocols keep $P$ and evaluate confidence values directly.
Hard-partition protocols first convert $P$ into a binary map $B$.
The fixed-threshold branch evaluates a specified operating point, commonly $P>\frac{1}{2}$.
The adaptive-threshold branch~\cite{Fmeasure} keeps $P\geq t_{\mathrm{adp}}$ with $t_{\mathrm{adp}}=\min\{2\bar P,1\}$, where $\bar P$ is the mean prediction value of each image.
The dynamic branch evaluates a threshold-indexed sequence $\{B_t\mid t\in\mathcal{T}\}$, allowing S4 scores such as \gls{Fbeta} and \gls{Em} to be reported as curves or summaries.
\cref{fig:thresholding_behaviors} illustrates the four behaviors on one prediction map.

\subsection{S2: Target Extraction}
\label{subsec:stage_entity_extraction}

The second stage determines what kind of target is extracted for later matching and scoring.
This stage has two ordered subdecisions in most protocols.
Target granularity is chosen first, either as a whole-map target or as independently parsed targets, and target type is then chosen among region, edge, and skeleton.
Most metrics in \cref{tab:stage_metric_mapping} keep the prediction and \gls{GT} as one image-level region target at this stage.
However, size-invariant metrics~\cite{SizeInvarianceVariants} adopt the separated-target region setting.
They parse connected components from the \gls{GT} mask, retain components that satisfy a minimum-area rule, and encode retained components by bounding-box regions~\cite{SizeInvarianceVariants}.
\gls{hIoU}~\cite{HierarchicalIoU} also uses separated region targets, but it parses connected components on both the thresholded prediction and the \gls{GT} so that predicted and annotated targets can be matched explicitly.
\gls{MSIoU}~\cite{MSIoU} adopts the whole-map edge setting and still treats the sample as one target instance, but changes the target type from region maps to whole-map Sobel edge maps before multiscale shrinking.
Surface-distance metrics also use whole-map edge targets, represented by boundary pixels extracted from the binary masks.
\gls{HCE}~\cite{HumanCorrectionEffortMeasure} is represented here by the whole-map skeleton setting, because its correction-effort mechanism relies on the skeleton of the whole \gls{GT} target.

\subsection{S3: Target Matching}
\label{subsec:stage_target_matching}

The third stage specifies how prediction-side and \gls{GT}-side elements are put into correspondence.
For the representative metrics analyzed here, the dominant mechanism is pixelwise correspondence.
The evaluated map may be a whole-map region, a target-aware local region, an edge map, or a skeleton-derived map, but the correspondence itself remains spatially aligned and pixelwise.
When S2 extracts separated targets, the size-invariant variants~\cite{SizeInvarianceVariants} analyzed here still use pixelwise correspondence inside each retained region.
Other target-level matching mechanisms can also be used at this stage.
They should be separated by the matching rule rather than by the pairwise score itself, because \gls{IoU}, Dice, centroid distance, or other measures can all serve as the affinity or cost used by different matching rules.
Threshold-only matching establishes correspondence directly from a pairwise acceptance condition, without a separate procedure for resolving competing candidates.
Globally optimized matching defines a global objective over all candidate pairs and chooses the assignment with minimum total cost or maximum total affinity.
Hungarian matching is a common one-to-one instance of this formulation.
The matching strategy of \gls{hIoU}~\cite{HierarchicalIoU} follows this target-level view with centroid distance as the cost.
These alternatives matter in fragmented, merged, or crowded predictions because unmatched predictions and unmatched annotations must then be handled explicitly.
Nearest-neighbor matching independently associates each element with the spatially closest element on the other side, allowing many-to-one correspondence.
Surface-distance metrics apply this rule to boundary points in one or both directions.
Implicit learned matching relates a candidate mask to an implicit reference encoded in an evaluator.
\gls{SegAE}~\cite{SegAE} uses this mechanism without explicit prediction-\gls{GT} pairs at evaluation time.

\glsunset{SIMAE}
\glsunset{BER}
\glsunset{ROC}
\begin{table}[t!]
  \centering
  \caption{Five-stage mechanism combinations for representative metric families.
  Entries are mechanism IDs from \cref{fig:metric_pipeline} with stage prefixes omitted, and / separates alternatives.
  \cref{tab:stage_metric_formulas} adds the compact formula of each family.}
  \label{tab:stage_metric_mapping}
  \adjustbox{max width=0.6\linewidth}{\begin{tabular}{@{}l c c c c c@{}}
\toprule
Metric &
\shortstack{(S1) Prediction\\Representation} &
\shortstack{(S2) Target\\Extraction} &
\shortstack{(S3) Target\\Matching} &
\shortstack{(S4) Score\\Computation} &
\shortstack{(S5) Metric\\Reporting} \\
\midrule
\glsxtrshort{MAE}                                                      & 1        & A1+B1    & 1        & A1       & 1        \\
\glsxtrshort{SIMAE}~\cite{SizeInvarianceVariants}                      & 1        & A2+B1    & 1        & A1       & 1        \\
Precision                                                      & 3/4      & A1+B1    & 1        & A1       & 1        \\
Recall                                                         & 3/4      & A1+B1    & 1        & A1       & 1        \\
\glsxtrshort{Fbeta}~\cite{Fmeasure}                                    & 3/4      & A1+B1    & 1        & A1       & 1/2/3    \\
SI-\glsxtrshort{Fbeta}~\cite{SizeInvarianceVariants}                   & 3/4      & A2+B1    & 1        & A1       & 1/2/3    \\
PR Curve                                                       & 4        & A1+B1    & 1        & A1       & 1        \\
\glsxtrshort{ROC} Curve                                                & 4        & A1+B1    & 1        & A1       & 1        \\
\glsxtrshort{AP}                                                       & 4        & A1+B1    & 1        & A1       & 4        \\
\glsxtrshort{IoU}                                                      & 2        & A1+B1    & 1        & A1       & 1        \\
Dice                                                           & 2        & A1+B1    & 1        & A1       & 1        \\
\glsxtrshort{OA}                                                       & 2        & A1+B1    & 1        & A1       & 1        \\
\glsxtrshort{BER}~\cite{TransparentObjectSegmentation,SurveyShadowDetection} & 2        & A1+B1    & 1        & A1       & 1        \\
\glsxtrshort{wFbeta}~\cite{wFmeasure}                                  & 1        & A1+B1    & 1        & A1+B2+B3 & 1        \\
\glsxtrshort{Sm}~\cite{Smeasure}                                       & 1        & A1+B1    & 1        & A2       & 1        \\
\glsxtrshort{Em}~\cite{Emeasure}                                       & 3/4      & A1+B1    & 1        & A1       & 1/2/3    \\
\glsxtrshort{MSIoU}~\cite{MSIoU}                                       & 2        & A1+B2    & 1        & A1+B1    & 1        \\
\glsxtrshort{HCE}~\cite{HumanCorrectionEffortMeasure}                  & 2        & A1+B3    & 1        & A3       & 1        \\
\glsxtrshort{Cm}~\cite{ContextMeasure}                                 & 1        & A1+B1    & 1        & A4+B3    & 1        \\
\glsxtrshort{wCm}~\cite{ContextMeasure}                          & 1        & A1+B1    & 1        & A4+B2+B3 & 1        \\
\glsxtrshort{hIoU}~\cite{HierarchicalIoU}                              & 2        & A2+B1    & 3        & A1       & 1        \\
\glsxtrshort{ASD}                                              & 2        & A1+B2    & 4        & A5       & 1        \\
\glsxtrshort{ASSD}                                             & 2        & A1+B2    & 4        & A5       & 1        \\
\glsxtrshort{HD}                                               & 2        & A1+B2    & 4        & A5       & 1        \\
\glsxtrshort{HD95}                                             & 2        & A1+B2    & 4        & A5       & 1        \\
\glsxtrshort{SegAE}~\cite{SegAE}                                       & 2        & A1+B1    & 5        & A6       & 1        \\
\glsxtrshort{FT}~\cite{FractionalThreshold}                            & 2        & A1+B1    & 1        & A1+B4    & 1        \\
\bottomrule
\end{tabular}%
}
\end{table}
\glsreset{BER}
\glsreset{ROC}

\subsection{S4: Score Computation}
\label{subsec:stage_similarity_cost}

The fourth stage applies the scoring rule using the correspondence or learned reference specified in S3, including any target, block, or scale combinations needed to form a complete score.
We distinguish six base scoring modes from four optional modifiers in \cref{fig:metric_pipeline}.
Base modes specify how quality is measured, while modifiers transform scoring evidence, its contributions, or the resulting quality values.
With the first three stages held fixed, a new base mode or modifier is introduced only for a substantive scoring difference that existing combinations do not capture.
For example, Dice and \gls{IoU} remain within the same mode because both derive quality from \gls{TP}, \gls{FP}, and \gls{FN}, despite using different formulas.

\parhead{Base Modes.}
Confusion-statistics scoring derives overlap, accuracy-like, and balance-oriented scores from a common set of target-background agreement and disagreement statistics.
It also includes \gls{MAE} for binary \gls{GT} under product-based soft confusion statistics, and \gls{Em}~\cite{Emeasure} for binary predictions, as shown in \cref{eq:soft_mae_confusion,eq:em_confusion_reduction}.
Structural-statistics scoring describes the internal structure of an evaluated entity, as in the foreground-background statistics and spatial block comparisons of \gls{Sm}~\cite{Smeasure}.
Edit cost measures the effort needed to repair a mask, as in \gls{HCE}~\cite{HumanCorrectionEffortMeasure}, which counts correction operations.
Bidirectional response scoring separately constructs prediction-to-\gls{GT} and \gls{GT}-to-prediction responses and combines their normalized values, as in \gls{Cm}~\cite{ContextMeasure}.
Spatial-distance statistics summarize distances between the spatial coordinates of corresponding elements, as in \gls{ASD} and \gls{HD}.
Implicit learned scoring estimates quality with a trained evaluator, as in \gls{SegAE}~\cite{SegAE}.

\parhead{Optional Modifiers.}
The modifiers describe computational roles and can be used individually or in combination.
Pyramid processing evaluates the same entities at multiple pyramid levels and combines the level-wise results.
\gls{MSIoU}~\cite{MSIoU} applies this mechanism to edge-recall scores.
Contribution weighting adjusts the weights assigned to scoring terms using information such as target geometry or image context.
The camouflage-aware variant of \gls{Cm}, for example, weights contextual responses using image-derived camouflage degree.
Contextual aggregation combines related evidence to construct the representation used for scoring.
\gls{wFbeta}~\cite{wFmeasure} combines contextual aggregation with contribution weighting.
Human calibration transforms base scores according to a reference derived from human performance or judgments.
\gls{FT}~\cite{FractionalThreshold} implements this modifier as a binary acceptance decision for each sample, using a threshold derived from human agreement.

\subsection{S5: Metric Reporting}
\label{subsec:stage_metric_reporting}

The fifth stage specifies how the score produced by the preceding stages is reported.
For soft grayscale, fixed-threshold, and adaptive-threshold protocols, the raw score data produced upstream is usually one scalar score.
For dynamic-threshold protocols, S1 produces a threshold-indexed prediction sequence, S4 computes a score at each threshold, and the raw data becomes a threshold-indexed score curve.
The raw-data mechanism unifies the original scalar value and the original score curve.
A threshold-indexed curve may then be summarized by threshold-wise mean, threshold-wise maximum, or threshold-wise \gls{AUC}.
A scalar score can be treated as a degenerate threshold curve on which mean, maximum, and normalized threshold-wise \gls{AUC} all reduce to the same scalar, so \cref{tab:stage_metric_mapping} marks scalar protocols by the more representative raw-data mechanism.
Curve reports include the precision-recall (PR) curve and the \gls{ROC} curve, and \gls{AP} summarizes the PR curve as a scalar.

\subsection{Dataset Aggregation}
\label{subsec:dataset_aggregation}

The five stages above define a metric on a single prediction-\gls{GT} pair.
Beyond the single sample, benchmark reporting further combines the results across samples.
This combination can enter the pipeline after different stages.
Aggregation after S3 sums the matched quantities across images before S4 computes the score, as in micro \gls{IoU} ($\frac{\sum_i I_i}{\sum_i U_i}$).
Aggregation after S4 averages the threshold-indexed score curves across images before S5 reporting, as in the maximum F-measure protocols of common toolkits.
Aggregation after S5 reports each image separately and then averages the reported scores, as in macro \gls{IoU} ($\frac{1}{N_{\mathrm{img}}}\sum_i \frac{I_i}{U_i}$).
The position does not affect the result when every later operation is linear in the aggregated quantity.
Raw data at a fixed threshold, threshold-wise mean, and threshold-wise \gls{AUC} are linear in the score curve, so aggregation after S4 and after S5 gives the same value.
Threshold-wise maximum and the ratios in confusion-statistics scoring are nonlinear, so the position can change scores and rankings.
We treat the aggregation position as a protocol choice that accompanies the five stages.

\section{Mapping Existing Metrics}
\label{sec:metric_definitions}

We use the five-stage framework to analyze representative metrics.
As shown in \cref{tab:stage_metric_mapping}, the mapping should be read as an index to metric protocols rather than as a catalogue of equations.
Each row specifies the stage path of one metric family, and each column compares one stage across metric families.
For example, the rows for \gls{MAE}, \gls{wFbeta}, \gls{Sm}, and \gls{Cm} all list S1-1, so these metrics keep a soft prediction before diverging at S4.
The rows for \gls{Fbeta}, \gls{IoU}, Dice, \gls{BER}, PR curves, and \gls{ROC} curves list S4-A1, so they share confusion statistics while using different score definitions and reporting modes.
In this reading, the compact formula identifies only part of the metric definition, whereas the stage identifiers specify the full evaluation path.
The compact formulas are listed in \cref{tab:stage_metric_formulas}, and complete metric formulas and the stage-level interpretation of each metric family are given in \cref{sup:metric_details}.

\section{Composing a Metric with the Framework}
\label{sec:metric_construction}

The stage vocabulary also supports constructing a metric from explicit evaluation requirements.
We consider four requirements, \ie, retaining confidence sensitivity, giving equal weight to \gls{GT} targets, including background errors across the entire image, and giving credit to spatial agreement at coarser resolutions.
The first requirement selects the soft grayscale map in S1.
The second selects separated-component regions in S2 and an equal mean of target-level scores in S4.
The bounding-box regions of the size-invariant \gls{Fbeta}~\cite{SizeInvarianceVariants} leave background errors outside all retained boxes unevaluated.
To satisfy the third requirement, we instead divide the entire image along the equal-distance midlines between \gls{GT} components.
The fourth requirement introduces pyramid processing (S4-B1) within each extracted region.
We retain pixelwise matching in S3 and soft \gls{IoU} as the local confusion-statistics score in S4.
We name the resulting metric \gls{ETMSIoU} after its equal target weights, pyramid processing, and soft scoring, and denote its score by $S_{\mathrm{ETMS}}$.
Its configuration follows S1-1 $\rightarrow$ S2-A2+B1 $\rightarrow$ S3-1 $\rightarrow$ S4-A1+B1 $\rightarrow$ S5-1 in \cref{fig:metric_pipeline}.
Let $C_1,\ldots,C_K$ be all four-connected \gls{GT} components, with no minimum-area filter.
We assign every image pixel to its nearest component by Euclidean distance, resolving ties by the smaller component label in raster order.
The resulting masks $M_k$ form a complete disjoint partition, and each contains its full \gls{GT} component with no foreground from another component.
We apply these masks before pyramid processing so that each target retains a separate prediction-\gls{GT} pair at every level.
At level $j$, let $\mathcal{D}_j$ take the maximum within each nonoverlapping cell of side $2^j$ pixels.
Maximum pooling keeps the confidence scale and gives binary occupancy for \gls{GT}.
The pooled maps are $P_{k,j}=\mathcal{D}_j(M_kP)$ and $G_{k,j}=\mathcal{D}_j(M_kG)$, giving the local score
\begin{equation}
  q_{k,j}=\frac{\sum_x P_{k,j}(x)G_{k,j}(x)}{\sum_x[P_{k,j}(x)+G_{k,j}(x)-P_{k,j}(x)G_{k,j}(x)]}.
  \label{eq:regional_scale_soft_iou}
\end{equation}
Following the scale-integration construction of \gls{MSIoU}~\cite{MSIoU}, we integrate these regional soft \gls{IoU} values across levels.
We use $J=10$ levels and the normalized scale coordinate $u_j=j/(J-1)$.
Let $q_k(u)$ linearly interpolate the samples $(u_j,q_{k,j})$.
The \gls{ETMSIoU} score is then
\begin{align}
  S_{\mathrm{ETMS}}
  = \frac{1}{K}\sum_{k=1}^{K}\int_0^1 q_k(u)\mathop{}d u
  = \frac{1}{2K(J-1)}\sum_{k=1}^{K}\sum_{j=0}^{J-2}(q_{k,j}+q_{k,j+1}).
  \label{eq:equal_target_multiscale_soft_iou}
\end{align}
Scale integration completes the local score before the equal-target average.
Each background prediction enters exactly one regional pyramid.
Pooling can merge nearby prediction and \gls{GT} pixels into the same coarse cell, allowing approximate spatial agreement to contribute.
The finest level retains the original regional soft \gls{IoU} with a positive weight.
With one \gls{GT} component, the score reduces to whole-map multiscale soft \gls{IoU}.
For empty \gls{GT}, we use the whole-map multiscale rule and assign zero to a zero denominator.

\section{Experiments}
\label{sec:experiments}

\subsection{Experimental Setup}
\label{subsec:experimental_setup}

The experiments test whether the five-stage decomposition explains how evaluation protocols behave.
Synthetic predictions and paired controls allow us to change one stage choice or one error property at a time while keeping every other factor fixed.
Each change in score or ranking can then be traced to a specific stage and mechanism.
We use two independently generated data sources, \ie, a shared benchmark for the protocol comparisons in \cref{subsec:protocol_sensitivity} and paired controls for the analyses in \cref{subsec:paired_controls}.
All their random streams derive from the fixed seed 42.
Construction details, implementation checks, aggregation conventions, and the results of additional controls appear in \cref{sup:controlled_setup}.

\subsection{Protocol Sensitivity Across Stages}
\label{subsec:protocol_sensitivity}

\begin{table}[t!]
  \centering
  \caption{Effects of evaluation protocols on prediction types and rankings.
  Baseline: Whole-Map \gls{IoU}, $P>0.5$, Both-Empty $0$, Raw Data, and Image Mean.
  All entries use the 300 \gls{GT} masks, and threshold-wise reports sweep $t=0,0.01,\ldots,1$.}
  \label{tab:controlled_cases}
  \adjustbox{max width=\linewidth}{\begin{tabular}{@{}lccccc@{}}
\toprule
Protocol & \shortstack[c]{Balanced\\Response} & \shortstack[c]{Low\\Confidence} & \shortstack[c]{Dilated\\Regions} & \shortstack[c]{Small-Component\\Omission} & \shortstack[c]{Pairwise Reversals\\ \vs Baseline} \\
\midrule
Baseline & 0.700 & 0.380 & 0.475 & 0.552 & 0 \\
\midrule
\multicolumn{6}{@{}l@{}}{\textit{Prediction Representation (S1)}} \\
Soft Grayscale Map & 0.185 & 0.201 & 0.182 & 0.254 & 3 \\
Fixed-Threshold Binary ($t=0.25$) & 0.596 & 0.792 & 0.427 & 0.508 & 3 \\
Fixed-Threshold Binary ($P\geq0.5$) & 0.702 & 0.384 & 0.469 & 0.552 & 0 \\
Fixed-Threshold Binary ($t=0.75$) & 0.468 & 0.000 & 0.531 & 0.456 & 2 \\
Adaptive-Threshold Binary & 0.586 & 0.470 & 0.435 & 0.455 & 2 \\
\midrule
\multicolumn{6}{@{}l@{}}{\textit{Target Extraction (S2, Whole Map + Region = Baseline)}} \\
Separated Component + Region & 0.696 & 0.379 & 0.572 & 0.503 & 1 \\
Whole Map + Edge & 0.514 & 0.229 & 0.006 & 0.489 & 1 \\
Whole Map + Skeleton & 0.438 & 0.184 & 0.445 & 0.461 & 2 \\
\midrule
\multicolumn{6}{@{}l@{}}{\textit{Target Matching (S3 on Separated Components, Pixelwise Matching = Baseline)}} \\
Threshold-only Matching ($\operatorname{IoU}>0.5$) & 0.687 & 0.265 & 0.354 & 0.511 & 0 \\
Globally Optimized Matching & 0.690 & 0.273 & 0.456 & 0.511 & 0 \\
Nearest-neighbor Matching & 0.690 & 0.274 & 0.456 & 0.511 & 0 \\
\midrule
\multicolumn{6}{@{}l@{}}{\textit{Score Computation (S4, Confusion Statistics (\gls{IoU}) = Baseline)}} \\
Structural Statistics ($\gls{Sm}$) & 0.757 & 0.586 & 0.615 & 0.673 & 0 \\
Contribution Weighting & 0.700 & 0.380 & 0.443 & 0.552 & 0 \\
Contextual Aggregation & 0.719 & 0.423 & 0.475 & 0.555 & 0 \\
Contribution Weighting + Contextual Aggregation & 0.718 & 0.423 & 0.443 & 0.555 & 0 \\
\midrule
\multicolumn{6}{@{}l@{}}{\textit{Empty-Case Handling}} \\
Both-Empty One & 0.900 & 0.580 & 0.675 & 0.752 & 0 \\
Both-Empty Ignored & 0.875 & 0.475 & 0.593 & 0.689 & 0 \\
\midrule
\multicolumn{6}{@{}l@{}}{\textit{Metric Reporting (S5, Raw Data = Baseline, Aggregation After S4)}} \\
Threshold-wise Mean & 0.438 & 0.333 & 0.370 & 0.403 & 0 \\
Threshold-wise Maximum & 0.716 & 0.792 & 0.535 & 0.555 & 3 \\
Threshold-wise AUC & 0.442 & 0.336 & 0.374 & 0.407 & 0 \\
\midrule
\multicolumn{6}{@{}l@{}}{\textit{Aggregation Position (Image Means After S4 or S5 = Baseline)}} \\
Micro \glsxtrshort{IoU} ($t=0.5$), Aggregation After S3 & 0.970 & 0.501 & 0.684 & 0.936 & 0 \\
Threshold-wise Maximum, Aggregation After S5 & 0.762 & 0.792 & 0.574 & 0.557 & 4 \\
\bottomrule
\end{tabular}
}
\end{table}

\parhead{Shared Benchmark.}
The protocol comparisons use 300 shared \gls{GT} masks at $160\times160$ resolution (\cref{sup:protocol_benchmark}).
The masks are divided equally into five groups, \ie, empty maps, small targets, large targets, multiple targets, and thin structures.
Each mask yields four prediction types.
\ptype{Balanced Response} uses a smoothed \gls{GT} mask with high foreground and low background confidence, while \ptype{Low Confidence} lowers foreground confidence and uses weaker smoothing for sharper boundaries.
\ptype{Dilated Regions} expands the \gls{GT} mask before smoothing, and \ptype{Small-Component Omission} removes components below an area cutoff and smooths the retained mask.
This gives 1,200 prediction-\gls{GT} pairs with floating-point confidence values in $[0,1]$.
We change one stage choice or the aggregation position at a time on this benchmark and compare the four prediction types with the whole-map \gls{IoU} baseline (\cref{tab:controlled_cases}).
\tcol{Pairwise Reversals \vs Baseline} counts the pairs among the four prediction types, six in total, whose order differs from \trow{Baseline}.

\parhead{Prediction Representation.}
The cutoff in S1 decides which prediction type ranks first.
\ptype{Balanced Response} ranks first at the baseline cutoff.
Lowering the cutoff favors \ptype{Low Confidence} by recovering target pixels within its sharper spatial support.
Raising the cutoff favors \ptype{Dilated Regions} by trimming excess foreground, and it empties every map of \ptype{Low Confidence}.
Adaptive thresholding follows the confidence scale of each map and moves \ptype{Low Confidence} above \ptype{Dilated Regions} and \ptype{Small-Component Omission}.
Soft scoring keeps the confidence values that thresholding discards, including background responses below the cutoff.
It places \ptype{Small-Component Omission} and \ptype{Low Confidence} above \ptype{Balanced Response}, whose background confidence is higher.
Changing the strict cutoff to $P\geq0.5$ alters only the absolute scores (\cref{tab:controlled_cases}).
All protocols score the stored predictions without per-image min-max normalization (\cref{sup:prediction_preprocessing}).

\parhead{Target Extraction.}
Separated-component region scoring averages local \gls{IoU} equally within retained \gls{GT} target boxes~\cite{SizeInvarianceVariants}, and it moves \ptype{Dilated Regions} above \ptype{Small-Component Omission} (\cref{sup:stage_protocols}).
A missed retained target affects its local box score, while false-positive pixels outside all target boxes do not enter the score.
Replacing regions with edges under whole-map granularity and pixelwise matching drops \ptype{Dilated Regions} to last, because its displaced boundaries rarely coincide with \gls{GT} boundaries.
Target type therefore changes the geometric property being evaluated even under the same overlap formula.

\begin{table}[t!]
  \centering
  \caption{Effects of merges, splits, missing or extra targets, and threshold-induced bridges under the four deterministic matching mechanisms.
    Threshold-only matching accepts a component pair when its \gls{IoU} exceeds $0.5$.
    Each entry is the mean over 300 sampled GT masks, with four-connected components.}
  \label{tab:matching_experiments}
  \adjustbox{max width=\linewidth}{\begin{tabular}{@{}lccccc@{}}
\toprule
Perturbation & \shortstack[c]{Components\\$B$ / $G$} & \shortstack[c]{Pixelwise\\Matching} & \shortstack[c]{Threshold-only\\Matching} & \shortstack[c]{Globally Optimized\\Matching} & \shortstack[c]{Nearest-neighbor\\Matching} \\
\midrule
\multicolumn{6}{@{}l@{}}{\textit{Entity Perturbations (Binary Predictions)}} \\
Perfect & 2/2 & 1.000 & 1.000 & 1.000 & 1.000 \\
Bridge (1 px) & 1/2 & 0.994 & 0.000 & 0.331 & 0.497 \\
Bridge (2 px) & 1/2 & 0.989 & 0.000 & 0.330 & 0.494 \\
Bridge (4 px) & 1/2 & 0.978 & 0.000 & 0.326 & 0.489 \\
Split (1 px) & 3/2 & 0.979 & 0.400 & 0.596 & 0.690 \\
Split (2 px) & 3/2 & 0.957 & 0.400 & 0.596 & 0.681 \\
Split (4 px) & 3/2 & 0.915 & 0.400 & 0.596 & 0.664 \\
Missing Component & 1/2 & 0.500 & 0.667 & 0.667 & 0.667 \\
Extra Component & 3/2 & 0.977 & 0.800 & 0.800 & 0.800 \\
\midrule
\multicolumn{6}{@{}l@{}}{\textit{Threshold-Induced Bridge (Target / Bridge / Background $0.9/0.4/0.1$, 2 px Bridge)}} \\
$P>0.25$ & 1/2 & 0.989 & 0.000 & 0.330 & 0.494 \\
$P>0.50$ & 2/2 & 1.000 & 1.000 & 1.000 & 1.000 \\
$P>0.90$ & 0/2 & 0.000 & 0.000 & 0.000 & 0.000 \\
\bottomrule
\end{tabular}
}
\end{table}

\parhead{Metric Reporting \& Aggregation Position.}
The S5 reports in \cref{tab:controlled_cases} follow common toolkits and average the threshold curves across images after S4.
Raw data, threshold-wise mean, and threshold-wise \gls{AUC} preserve the baseline ranking.
Threshold-wise maximum moves \ptype{Low Confidence} from last to first.
Together with the lowest mean and \gls{AUC}, this high maximum shows that strong overlap for \ptype{Low Confidence} is confined to a narrow threshold range.
Moving the aggregation after S5 lets every image select its own threshold and also raises \ptype{Dilated Regions} above \ptype{Small-Component Omission}.
Micro aggregation after S3 preserves the ranking but raises \ptype{Small-Component Omission} close to \ptype{Balanced Response}, because large targets dominate the summed pixel counts and the omitted small components barely affect them.
A reported score therefore needs a stated aggregation position whenever a nonlinear operation follows it.

\subsection{Controlled Analyses of Stage Mechanisms}
\label{subsec:paired_controls}

\parhead{Paired Controls.}
The analyses here use paired controls whose \gls{GT} masks are generated separately.
Each paired control builds its \gls{GT} from simple shapes, such as two squares, a thin line, or a rectangle, so that one error property can change while the others stay fixed.
The control repeats this construction on 300 \gls{GT} masks with randomly sampled target positions and sizes, using the $160\times160$ resolution and random seed of the shared benchmark, and all compared predictions are built on the same 300 masks.

\subsubsection{Tracing Responses to Stages}
\label{subsubsec:mechanism_attribution}

\glsunset{ASSD}\glsunset{HD95}
\begin{figure}[t!]
  \centering
  \includegraphics[width=\linewidth]{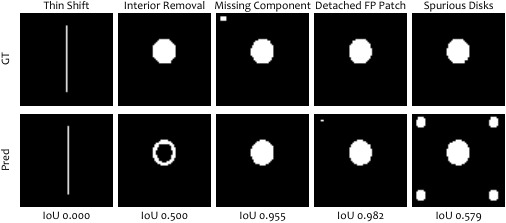}
  \caption{Representative error types used for metric-response analysis.
    Each column shows a single sample, with its whole-map \gls{IoU} printed below.
    Each case changes one error property, \ie, displacement, region loss, a missing component, or detached false positives.}
  \label{fig:controlled_perturbations}
  \vspace{0.5em}
  \captionof{table}{Responses of overlap, structural, edge, and surface-distance metrics to the error types in \cref{fig:controlled_perturbations}.
  Each entry is the mean over 300 sampled GT masks for binary predictions, with distances in pixels.}
  \label{tab:failure_experiments}
  \adjustbox{max width=\linewidth}{\begin{tabular}{@{}lrrrrrrrrrrr@{\hspace{1.5em}}@{}}
\toprule
Error Type & \multicolumn{1}{c}{\diaghead{\gls{IoU} $\uparrow$}} & \multicolumn{1}{c}{\diaghead{\gls{MAE} $\downarrow$}} & \multicolumn{1}{c}{\diaghead{\glsxtrshort{wFbeta} $\uparrow$}} & \multicolumn{1}{c}{\diaghead{\glsxtrshort{Sm} $\uparrow$}} & \multicolumn{1}{c}{\diaghead{\glsxtrshort{Em} $\uparrow$}} & \multicolumn{1}{c}{\diaghead{\glsxtrshort{MSIoU}$_{J=1}$ $\uparrow$}} & \multicolumn{1}{c}{\diaghead{\glsxtrshort{MSIoU}$_{J=10}$ $\uparrow$}} & \multicolumn{1}{c}{\diaghead{\gls{ASD} $\downarrow$}} & \multicolumn{1}{c}{\diaghead{\gls{ASSD} $\downarrow$}} & \multicolumn{1}{c}{\diaghead{\gls{HD} $\downarrow$}} & \multicolumn{1}{c}{\diaghead{\gls{HD95} $\downarrow$}} \\
\midrule
Thin Shift (1 px) & 0.500 & 0.009 & 0.740 & 0.748 & 0.993 & 0.514 & 0.933 & 0.983 & 0.983 & 1.000 & 1.000 \\
Thin Shift (3 px) & 0.000 & 0.027 & 0.000 & 0.464 & 0.979 & 0.497 & 0.850 & 2.000 & 2.000 & 3.000 & 3.000 \\
Thin Shift (6 px) & 0.000 & 0.027 & 0.000 & 0.464 & 0.979 & 0.002 & 0.663 & 5.000 & 5.000 & 6.000 & 6.000 \\
Interior Removal (50\% of $G$) & 0.500 & 0.049 & 0.705 & 0.741 & 0.789 & 1.000 & 1.000 & 2.780 & 1.763 & 7.087 & 6.984 \\
Missing $8\times8$ Component of $G$ & 0.973 & 0.002 & 0.987 & 0.988 & 0.998 & 0.867 & 0.832 & 0.000 & 5.009 & 70.074 & 66.442 \\
Detached FP Pixel & 1.000 & 0.000 & 1.000 & 0.998 & 1.000 & 1.000 & 1.000 & 0.382 & 0.192 & 64.460 & 0.000 \\
Detached $5\times5$ FP Patch & 0.989 & 0.001 & 0.989 & 0.990 & 0.999 & 1.000 & 1.000 & 5.598 & 2.931 & 67.266 & 63.699 \\
Four Spurious Disks ($r=9$ px) & 0.706 & 0.039 & 0.706 & 0.838 & 0.915 & 1.000 & 1.000 & 30.550 & 20.852 & 76.504 & 72.159 \\
\bottomrule
\end{tabular}
}
\end{figure}

\begin{figure}[t!]
  \centering
  \includegraphics[width=\linewidth]{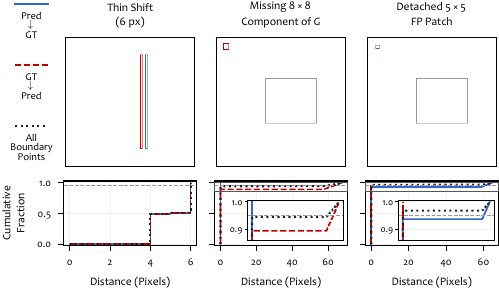}
  \caption{Surface-distance controls and distance distributions.
  \textit{Upper}: \textcolor[HTML]{C00000}{Red}, \textcolor[HTML]{3366CC}{blue}, and \textcolor[HTML]{999999}{gray} mark GT, prediction, and coincident boundary points.
  \textit{Lower}: Cumulative distance distributions, with the \textcolor[HTML]{999999}{gray} dashed line at $0.95$ and insets enlarging the framed bands.
  A thin shift affects both directions equally, a missing component affects GT~$\rightarrow$~Pred, and a detached false-positive patch reaches the 95th percentile for Pred~$\rightarrow$~GT but not over all boundary points.}
  \label{fig:surface_distance_controls}
\end{figure}

\parhead{Matching \& Error-Type Controls.}
The matching controls in \cref{tab:matching_experiments} show that the matching mechanisms of S3 score merged and split targets differently, and that a threshold change in S1 alters the entities S2 and S3 receive (\cref{sup:matching_constructions}).
Error-type controls compare region, edge, and surface-distance responses (\cref{tab:failure_experiments}) to the displacement, region loss, missing components, and detached false positives illustrated in \cref{fig:controlled_perturbations}.
\cref{fig:surface_distance_controls} further shows the distance distributions behind the surface-distance responses (\cref{sup:error_type_constructions}).

\begin{table}[t!]
  \centering
  \caption{Responses to the layouts (a)--(d) in \cref{fig:scoring_controls}.
  Each entry is the mean over 300 sampled GT masks for binary predictions with $\beta^2=1$.}
  \label{tab:fixed_count_controls}
  \adjustbox{max width=0.5\linewidth}{\begin{tabular}{@{}lrrrr@{}}
\toprule
Metric & (a) & (b) & (c) & (d) \\
\midrule
\gls{IoU} $\uparrow$ & 0.275 & 0.275 & 0.275 & 0.275 \\
\gls{MAE} $\downarrow$ & 0.290 & 0.290 & 0.290 & 0.290 \\
\glsxtrshort{Em} $\uparrow$ & 0.564 & 0.564 & 0.564 & 0.564 \\
\glsxtrshort{Sm} $\uparrow$ & 0.468 & 0.468 & 0.465 & 0.465 \\
\gls{HCE} $\downarrow$ & 13.000 & 10.000 & 13.000 & 10.000 \\
No Modifiers (\glsxtrshort{Fbeta}) $\uparrow$ & 0.431 & 0.431 & 0.431 & 0.431 \\
Contribution Weighting $\uparrow$ & 0.406 & 0.400 & 0.406 & 0.400 \\
Contextual Aggregation $\uparrow$ & 0.452 & 0.452 & 0.471 & 0.471 \\
Both Modifiers (\glsxtrshort{wFbeta}) $\uparrow$ & 0.425 & 0.419 & 0.444 & 0.438 \\
\bottomrule
\end{tabular}
}
\end{table}

\begin{figure}[t!]
  \centering
  \includegraphics[width=\linewidth]{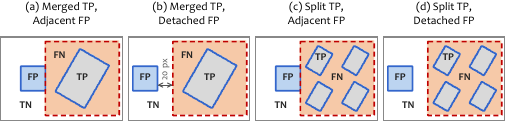}
  \caption{Error layouts at fixed confusion counts.
  \textcolor[HTML]{C00000}{Red} dashed and \textcolor[HTML]{3366CC}{blue} solid contours denote GT and predictions.
  All four predictions share the same GT and identical TP, FP, FN, and TN counts.}
  \label{fig:scoring_controls}
\end{figure}

\parhead{Confusion, Structure, \& Edit Cost (S4-A1--A3).}
To separate sensitivity to the spatial layout of errors from sensitivity to their area, we move fixed TP and FP areas as in \cref{fig:scoring_controls} while preserving the \gls{GT} and all four confusion counts (\cref{sup:scoring_controls}).
In \cref{tab:fixed_count_controls}, \gls{IoU}, \gls{MAE}, and binary \gls{Em} remain unchanged across layouts (a)--(d), which agrees with their reduction to confusion statistics (S4-A1) in \cref{eq:soft_mae_confusion,eq:em_confusion_reduction}.
Splitting the TP area slightly lowers \gls{Sm} through its block statistics (S4-A2), while the FP placement leaves \gls{Sm} unchanged in this construction.
The edit cost of \gls{HCE} (S4-A3) is the same for the merged and split TP areas, because both leave one FN region attached to the \gls{GT} boundary.
It assigns a higher cost to the adjacent FP block than to the detached one, because an attached error is corrected by moving boundary control points whereas a detached error is removed as one region.

\parhead{Scoring Modifiers (S4-B2--B3).}
The modifier ablation of \gls{wFbeta} separates contribution weighting from contextual aggregation on the fixed-count layouts (\cref{tab:fixed_count_controls}).
The ablation switches them independently on the binary error map $E=|B-G|$, following the effective-error construction in \cref{sup:stage_protocols}, and combines precision and recall with $\beta^2=1$.
Contribution weighting (S4-B2) lowers the score when the FP block is detached, because distance-based weights grow away from the \gls{GT}, and it ignores whether the TP area is merged or split.
Contextual aggregation (S4-B3) raises the score when the TP area is split, because more FN pixels then have correctly predicted foreground neighbors, and it ignores the FP placement.
The complete \gls{wFbeta} responds to both factors.
Contribution weighting and \gls{HCE} therefore order the two FP placements in opposite directions, as one measures distance from the target and the other counts correction operations.
RGB background controls further isolate the camouflage-derived contribution weighting of \glsxtrshort{wCm} (\cref{sup:context_controls}).

\subsubsection{Evaluating the Composed Metric}
\label{subsubsec:composed_metric_evaluation}

\begin{table}[t!]
  \centering
  \caption{Effects of successive configuration changes from whole-map hard \gls{IoU} to the constructed metric, \ie, \gls{ETMSIoU}.
  Prediction rows give mean scores over 300 sampled GT layouts.}
  \label{tab:composition_experiments}
  \adjustbox{max width=\linewidth}{\begin{tabular}{@{}lrrrr@{}}
\toprule
Prediction & \shortstack[c]{Whole-Map\\Hard \gls{IoU}\\($P>0.5$)} & \shortstack[c]{+ Hard\\$\rightarrow$\\Soft} & \shortstack[c]{+ Whole-Map $\rightarrow$\\Component Regions\\(Equal Weights)} & \shortstack[c]{+ Pyramid\\Processing\\(\gls{ETMSIoU})} \\
\midrule
\multicolumn{5}{@{}l@{}}{\textit{Confidence ($P=0.75G$ vs. $0.25G$)}} \\
Higher Confidence & 1.000 & 0.750 & 0.750 & 0.750 \\
Lower Confidence & 0.000 & 0.250 & 0.250 & 0.250 \\
Pairs Meeting the Requirement (of 300) & 300 & 300 & 300 & 300 \\
\midrule
\multicolumn{5}{@{}l@{}}{\textit{Target Coverage (Equal Pixel Errors, $A_{\mathrm{large}}/A_{\mathrm{small}}=5$)}} \\
Partial Misses on Both Targets & 0.833 & 0.833 & 0.700 & 0.876 \\
Small Target Omitted & 0.833 & 0.833 & 0.500 & 0.500 \\
Pairs Meeting the Requirement (of 300) & 0 & 0 & 300 & 300 \\
\midrule
\multicolumn{5}{@{}l@{}}{\textit{Background FP ($\mathrm{FP}/|G|=0.25$ vs. $1$)}} \\
Fewer Background FP & 0.800 & 0.800 & 0.709 & 0.663 \\
More Background FP & 0.500 & 0.500 & 0.459 & 0.621 \\
Pairs Meeting the Requirement (of 300) & 300 & 300 & 300 & 300 \\
\midrule
\multicolumn{5}{@{}l@{}}{\textit{Displacement (3 vs. 24 Pixels, No Pixel Overlap)}} \\
Nearby Predictions & 0.000 & 0.000 & 0.000 & 0.661 \\
Distant Predictions & 0.000 & 0.000 & 0.000 & 0.316 \\
Pairs Meeting the Requirement (of 300) & 0 & 0 & 0 & 300 \\
\midrule
\multicolumn{5}{@{}l@{}}{\textit{Combined Errors (Target Coverage, Displacement, and Background FP)}} \\
Both Targets With a Small Shift & 0.749 & 0.544 & 0.455 & 0.593 \\
Target Omission and Remote FP & 0.714 & 0.714 & 0.459 & 0.545 \\
Pairs Meeting the Requirement (of 300) & 162 & 0 & 162 & 277 \\
\bottomrule
\end{tabular}
}
\end{table}

To examine whether \gls{ETMSIoU} (\cref{eq:equal_target_multiscale_soft_iou}) meets the requirements that define it in \cref{sec:metric_construction} and which mechanism serves each requirement, we add its mechanisms one at a time to whole-map hard \gls{IoU} (\cref{tab:composition_experiments}).
The three additions are soft grayscale values, complete component-distance regions with equal target weights, and independent regional pyramids with scale integration.
We derive five paired tests from the four requirements in \cref{sec:metric_construction} and fix the preferred prediction of every pair before scoring.
Every pair is built so that its requirements, with stated priorities, decide which prediction should score higher.
All tests use 300 sampled \gls{GT} layouts, each with a small rectangle and a five times wider large rectangle.
Each group of \cref{tab:composition_experiments} lists the preferred prediction first, with the pair constructions in \cref{sup:composition_constructions}.
Within each test, the last row counts the pairs in which the preferred prediction scores strictly higher, and a tie counts as a failure.

All four configurations favor the higher-confidence prediction, but only the soft variants retain the confidence level instead of switching between omission and exact recovery at the cutoff.
Every configuration favors fewer background errors.
Both whole-map variants tie the pair in \tgroup{Target Coverage} because both predictions have identical confusion counts, and equal target averaging resolves the tie in favor of retaining both targets.
All original-resolution configurations tie the pair in \tgroup{Displacement} because neither prediction overlaps \gls{GT}, and regional pyramids distinguish nearby from distant predictions through agreement at coarser levels.
In \tgroup{Combined Errors}, whole-map soft scoring favors the prediction with an omitted target, and target averaging alone recovers the stated preference in 162 of the 300 pairs.
The full construction recovers it in 277 pairs (\cref{sup:composition_constructions}).

The paired tests are derived from the same requirements that define \gls{ETMSIoU}, and the four single-requirement tests hold for every sampled mask.
The selected stage mechanisms therefore implement these requirements, and the intermediate configurations show which requirement each mechanism serves.
The combined exceptions show that the response to spatial errors still depends on region assignment and grid alignment.

\section{Discussion}
\label{sec:discussion}

\parhead{Benchmark Reporting.}
Reproducibility is part of the metric definition.
A benchmark should document the prediction preprocessing, prediction representation, threshold policy, target entity, matching rule, score formula, empty-case convention, reporting rule, and aggregation position, because each can change reported scores (\cref{subsec:protocol_sensitivity}).
When these details are missing, reported numbers from different papers are hard to compare, even if the metric names are the same.
A benchmark should pair general agreement scores with task-aware measures when the task depends on fine boundaries, small targets, correction effort, or contextual ambiguity.

\parhead{Open Design Space.}
The five-stage view also shows that current metrics cover only part of the possible design space.
The option sets shown in \cref{fig:metric_pipeline} already imply many combinations that are rarely explored.
\gls{ETMSIoU} realizes one of them by keeping a soft prediction in S1 and extracting connected targets in S2 (\cref{sec:metric_construction}).
A metric could also extract boundary entities from a soft prediction, combine explicit target matching in S3 with structure-aware or effort-aware scoring in S4, or report distributions over target size, difficulty, or operating point in S5 alongside a single mean.
Genuinely new metric mechanisms can also be organized by the same five stages (\cref{sup:extended_discussion}).
Future metric proposals should explain which visual error they make visible, how that error is counted, and why the protocol matches the application.

\section{Conclusion}

This paper organizes binary target segmentation metrics into a five-stage framework.
The framework makes the protocol choices of existing metrics explicit, so that formulas, procedural choices, and reported scores can be discussed in the same language.
Mapped onto these stages, different metrics appear as different compositions of shared mechanisms, each serving a different evaluation need.
The protocol comparisons show that a single stage choice can reorder predictions under the same formula, and the controlled analyses trace metric responses to specific stage mechanisms.
The constructed \gls{ETMSIoU} and paired tests illustrate how the same vocabulary supports metric construction and verification.
Metric formulas, protocol choices, procedural details, and reported values must be stated and interpreted together.

\newpage
\appendix

\startcontents[appendix]
\section*{Appendix Contents}
\begingroup
\small
\titlecontents{section}[1.8em]{\addvspace{3pt}\bfseries}
  {\contentslabel[\thecontentslabel.]{1.8em}}{}{\hfill\contentspage}
\titlecontents{subsection}[4.4em]{}
  {\contentslabel[\thecontentslabel.]{2.6em}}{}{\titlerule*[0.6pc]{.}\contentspage}
\titlecontents{subsubsection}[7.8em]{}
  {\contentslabel[\thecontentslabel.]{3.4em}}{}{\titlerule*[0.6pc]{.}\contentspage}
\let\glsxtrfull\glspdffmtfull
\printcontents[appendix]{}{1}{\setcounter{tocdepth}{3}}
\endgroup

\newpage

\glsunsetall
\begin{table}[t!]
  \centering
  \caption{Five-stage mechanism combinations and compact formulas for representative metric families.
  Entries are mechanism IDs from \cref{fig:metric_pipeline} with stage prefixes omitted, and / separates alternatives.
  The boundary symbols $\partial B$ and $\partial G$ denote the prediction and \gls{GT} boundaries.}
  \label{tab:stage_metric_formulas}
  \resizebox{\linewidth}{!}{\begin{tabular}{@{}l c c c c c c@{}}
\toprule
Metric & Formula &
\shortstack{(S1) Prediction\\Representation} &
\shortstack{(S2) Target\\Extraction} &
\shortstack{(S3) Target\\Matching} &
\shortstack{(S4) Score Computation\\Base + Modifiers} &
\shortstack{(S5) Metric\\Reporting} \\
\midrule
\glsxtrfull{MAE}                                                                                                                               &
$\frac{1}{HW}\sum_x |P(x)-G(x)|$                                                                                                          &
1                                                                                                                                    & A1+B1 & 1 & A1 & 1                     \\
\glsxtrfull{SIMAE}~\cite{SizeInvarianceVariants}                                                                                               &
$\operatorname{Agg}_{k}\operatorname{MAE}_{k}$                                                                                          &
1                                                                                                                                    & A2+B1 & 1 & A1 & 1                     \\
Precision                                                                                                                               &
$\frac{\operatorname{TP}}{\operatorname{TP}+\operatorname{FP}}$                                                                         &
3/4                                                                                                                               & A1+B1 & 1 & A1 & 1                     \\
Recall                                                                                                                                  &
$\frac{\operatorname{TP}}{\operatorname{TP}+\operatorname{FN}}$                                                                         &
3/4                                                                                                                               & A1+B1 & 1 & A1 & 1                     \\
\glsxtrfull{Fbeta}~\cite{Fmeasure}                                                                                                                            &
$\frac{(1+\beta^2)\operatorname{Precision}\operatorname{Recall}}{\beta^2\operatorname{Precision}+\operatorname{Recall}}$               &
3/4                                                                                                                               & A1+B1 & 1 & A1 & 1/2/3           \\
Size-Invariant \glsxtrlong{Fbeta} (SI-\glsxtrshort{Fbeta})~\cite{SizeInvarianceVariants}                                                                                           &
$\operatorname{Agg}_{k}F_{\beta,k}$                                                                                                     &
3/4                                                                                                                               & A2+B1 & 1 & A1 & 1/2/3           \\
Precision-Recall (PR) Curve                                                                                                                                &
$\{(\operatorname{Precision}_{t},\operatorname{Recall}_{t})\mid t\in\mathcal{T}\}$                                                       &
4                                                                                                                                    & A1+B1 & 1 & A1 & 1                     \\
\glsxtrfull{ROC} Curve                                                                                                                         &
$\{(\operatorname{FPR}_{t},\operatorname{TPR}_{t})\mid t\in\mathcal{T}\}$                                                                &
4                                                                                                                                    & A1+B1 & 1 & A1 & 1                     \\
\glsxtrfull{AP}                                                                                                                                &
$\int_{\mathcal{T}}\operatorname{Precision}(t)\mathop{}d\operatorname{Recall}(t)$                                                       &
4                                                                                                                                    & A1+B1 & 1 & A1 & 4                     \\
\glsxtrfull{IoU}                                                                                                                               &
$\frac{\operatorname{TP}}{\operatorname{TP}+\operatorname{FP}+\operatorname{FN}}$                                                       &
2                                                                                                                                    & A1+B1 & 1 & A1 & 1                     \\
Dice                                                                                                                                    &
$\frac{2\operatorname{TP}}{2\operatorname{TP}+\operatorname{FP}+\operatorname{FN}}$                                                     &
2                                                                                                                                    & A1+B1 & 1 & A1 & 1                     \\
\glsxtrfull{OA}                                                                                                                                &
$\frac{\operatorname{TP}+\operatorname{TN}}{\operatorname{TP}+\operatorname{FP}+\operatorname{TN}+\operatorname{FN}}$                  &
2                                                                                                                                    & A1+B1 & 1 & A1 & 1                     \\
\glsxtrfull{BER}~\cite{TransparentObjectSegmentation,SurveyShadowDetection}                                                                    &
$\frac{\operatorname{FNR}+\operatorname{FPR}}{2}$                                                                                       &
2                                                                                                                                    & A1+B1 & 1 & A1 & 1                     \\

\glsxtrfull{wFbeta}~\cite{wFmeasure}                                                                                                  &
$\frac{(1+\beta^2)\operatorname{Precision}^{w}\operatorname{Recall}^{w}}{\beta^2\operatorname{Precision}^{w}+\operatorname{Recall}^{w}}$ &
1                                                                                                                                    & A1+B1 & 1 & A1+B2+B3 & 1                     \\
\glsxtrfull{Sm}~\cite{Smeasure}                                                                                                                &
$\alpha S_o+(1-\alpha)S_r$                                                                                                             &
1                                                                                                                                    & A1+B1 & 1 & A2 & 1                     \\
\glsxtrfull{Em}~\cite{Emeasure}                                                                                                                &
$\frac{1}{HW}\sum_x\frac{[1+a(x)]^2}{4}$                                                                                                &
3/4                                                                                                                               & A1+B1 & 1 & A1 & 1/2/3           \\
\glsxtrfull{MSIoU}~\cite{MSIoU}                                                                                                                &
$\int_{0}^{1} r(u)\mathop{}d u$                                                                                                                &
2                                                                                                                                    & A1+B2 & 1 & A1+B1 & 1                     \\
\glsxtrfull{HCE}~\cite{HumanCorrectionEffortMeasure}                                                                                           &
$N_{\mathrm{poly}}^{\mathrm{FP}}+N_{\mathrm{ind}}^{\mathrm{FP}}+N_{\mathrm{poly}}^{\mathrm{FN}}+N_{\mathrm{ind}}^{\mathrm{FN}}$          &
2                                                                                                                                    & A1+B3 & 1 & A3 & 1                     \\
\glsxtrfull{Cm}~\cite{ContextMeasure}                                                                                                          &
$\frac{(1+\beta^2)F_mR_m}{\beta^2F_m+R_m}$                                                                                              &
1                                                                                                                                    & A1+B1 & 1 & A4+B3 & 1                     \\
\glsxtrfull{wCm}~\cite{ContextMeasure} &
$\frac{(1+\beta^2)F_mR_w}{\beta^2F_m+R_w}$ &
1 & A1+B1 & 1 & A4+B2+B3 & 1 \\

\glsxtrfull{hIoU}~\cite{HierarchicalIoU}                                                                                                       &
$\frac{\sum_{\pi\in\mathcal{M}}q_{\pi}}{\mathrm{TP}_{tgt}+\mathrm{FP}_{tgt}+\mathrm{FN}_{tgt}}$                                          &
2                                                                                                                                    & A2+B1 & 3 & A1 & 1                     \\
\glsxtrfull{ASD} &
$\displaystyle\frac{1}{|\partial B|}\sum_{x\in\partial B}\min_{y\in\partial G}\|x-y\|_2$ &
2 & A1+B2 & 4 & A5 & 1 \\
\glsxtrfull{ASSD} &
$\frac{\displaystyle\sum_{x\in\partial B}\min_{y\in\partial G}\|x-y\|_2+\displaystyle\sum_{y\in\partial G}\min_{x\in\partial B}\|y-x\|_2}{|\partial B|+|\partial G|}$ &
2 & A1+B2 & 4 & A5 & 1 \\
\glsxtrfull{HD} &
$\displaystyle\max\!\left\{\max_{x\in\partial B}\min_{y\in\partial G}\|x-y\|_2,\max_{y\in\partial G}\min_{x\in\partial B}\|y-x\|_2\right\}$ &
2 & A1+B2 & 4 & A5 & 1 \\
\glsxtrfull{HD95} &
$\displaystyle\max\!\left\{\underset{x\in\partial B}{\operatorname{percentile}_{95}}\!\left(\min_{y\in\partial G}\|x-y\|_2\right),\underset{y\in\partial G}{\operatorname{percentile}_{95}}\!\left(\min_{x\in\partial B}\|y-x\|_2\right)\right\}$ &
2 & A1+B2 & 4 & A5 & 1 \\
\glsxtrfull{SegAE}~\cite{SegAE} &
$\mathcal{Q}_{\theta}(I,B,c)$ &
2 & A1+B1 & 5 & A6 & 1 \\
\glsxtrfull{FT}~\cite{FractionalThreshold} &
$\mathbf{1}[q_i \geq q_{\mathrm{hum}}]$ &
2 & A1+B1 & 1 & A1+B4 & 1 \\
\bottomrule
\end{tabular}%
}
\end{table}
\glsresetall

\section{Metric Formulas}
\label{sup:metric_details}

This section gives detailed formulas, computational conventions, and the stage-level interpretation of each metric family for the metrics mapped in \cref{tab:stage_metric_mapping}.
\cref{tab:stage_metric_formulas} first gives an overview that extends this mapping with the compact formula of each metric family.
The mapping also separates three kinds of metric variation that are often mixed together.
If two metrics differ only at S5, their raw score sequence may be the same even though the reported scalar differs.
If they differ at S2 or S3, they may evaluate different target entities and expose different failures.
If they differ at S4, they usually retain the same upstream evidence but ask a different quality question.

\subsection{Confusion-Matrix Metrics}

Confusion-matrix metrics form a tightly connected family because they derive scores from target-background agreement and disagreement statistics.
This shared-statistics formulation lets absolute-error, overlap, accuracy-like, balance-oriented, and curve-based scores reuse the same upstream quantities while differing mainly in the S4 formula and S5 reporting mechanism.
For example, \gls{MAE} measures soft disagreement under binary \gls{GT}, \gls{IoU} and Dice focus on target overlap, \gls{OA} measures whole-image correctness, \gls{BER}~\cite{TransparentObjectSegmentation,SurveyShadowDetection} balances target and background recognition, and \gls{ROC} or precision-recall based reports summarize threshold-indexed rate trade-offs.
Under a common fixed-threshold image protocol, overlap, accuracy-like, and balance-oriented confusion-matrix metrics usually produce scalar values represented as raw data, whereas dynamic-threshold protocols retain or summarize curves.
\gls{Em}, \gls{wFbeta}, \gls{MSIoU}, and \gls{hIoU} reuse these statistics with an alignment-based score, weighted error evidence, edge targets, and explicit target matching, respectively.

\subsubsection{Confusion Counts and Masses}

The shared quantities are most commonly \gls{TP}, \gls{FP}, \gls{TN}, and \gls{FN}.
S1 determines whether these quantities are computed as pixel counts from binary predictions or as soft masses from grayscale predictions.

\begin{figure}[t]
  \centering
  \begin{subfigure}{0.155\linewidth}
    \centering
    \includegraphics[width=\linewidth]{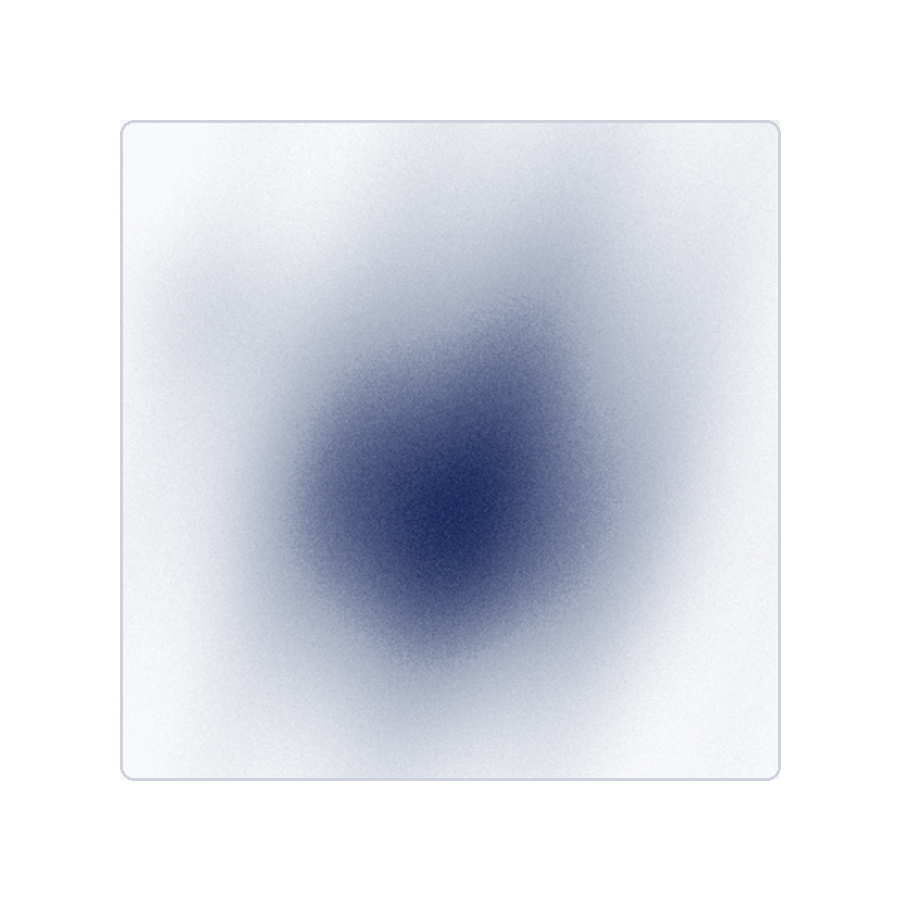}
    \caption{Soft}
    \label{fig:thresholding_soft}
  \end{subfigure}
  \hfill
  \begin{subfigure}{0.155\linewidth}
    \centering
    \includegraphics[width=\linewidth]{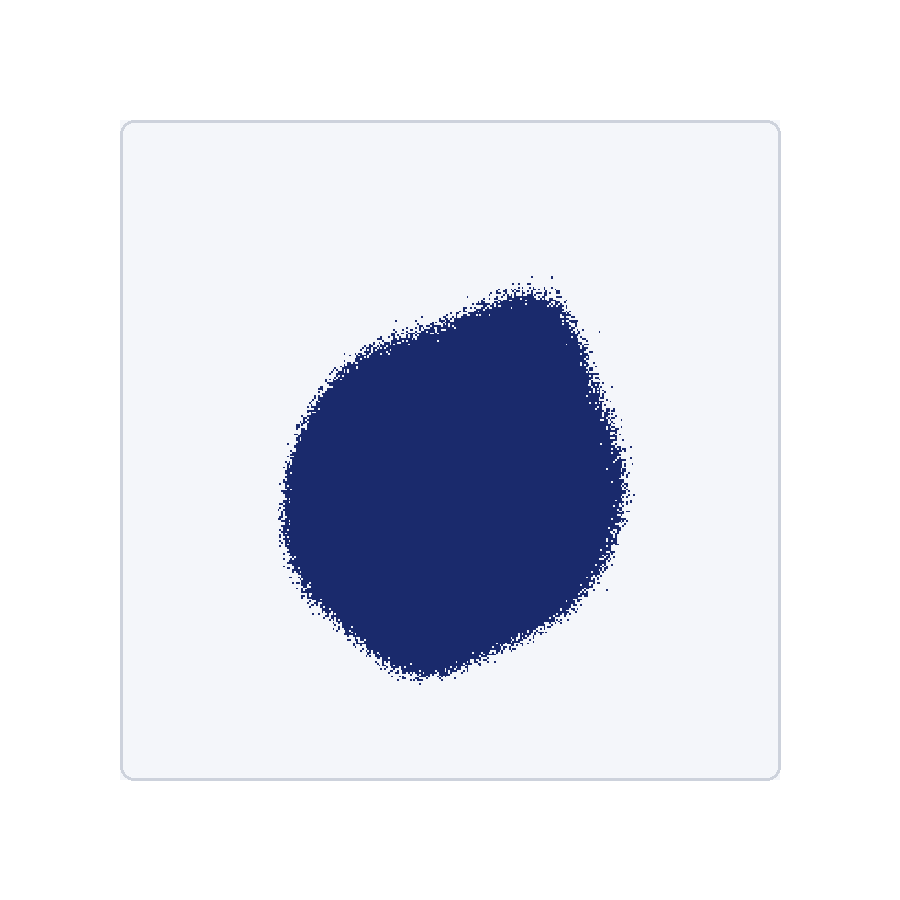}
    \caption{Fixed}
    \label{fig:thresholding_fixed}
  \end{subfigure}
  \hfill
  \begin{subfigure}{0.155\linewidth}
    \centering
    \includegraphics[width=\linewidth]{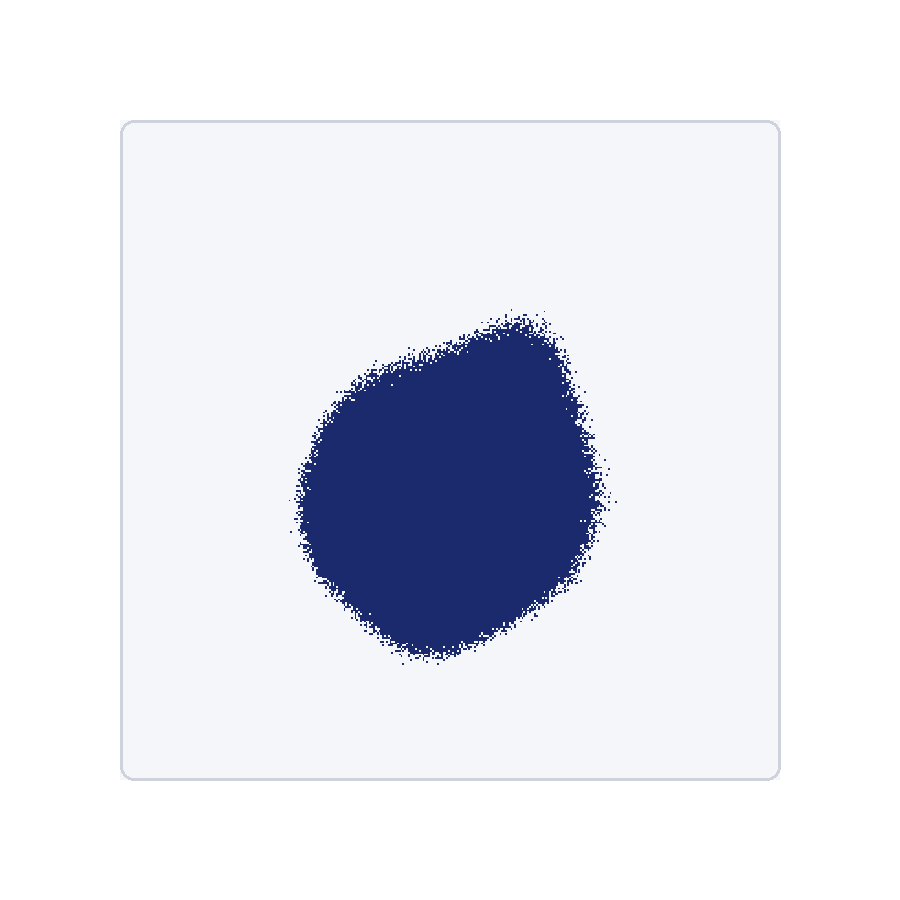}
    \caption{Adaptive}
    \label{fig:thresholding_adaptive}
  \end{subfigure}
  \hfill
  \begin{subfigure}{0.465\linewidth}
    \centering
    \includegraphics[width=\linewidth]{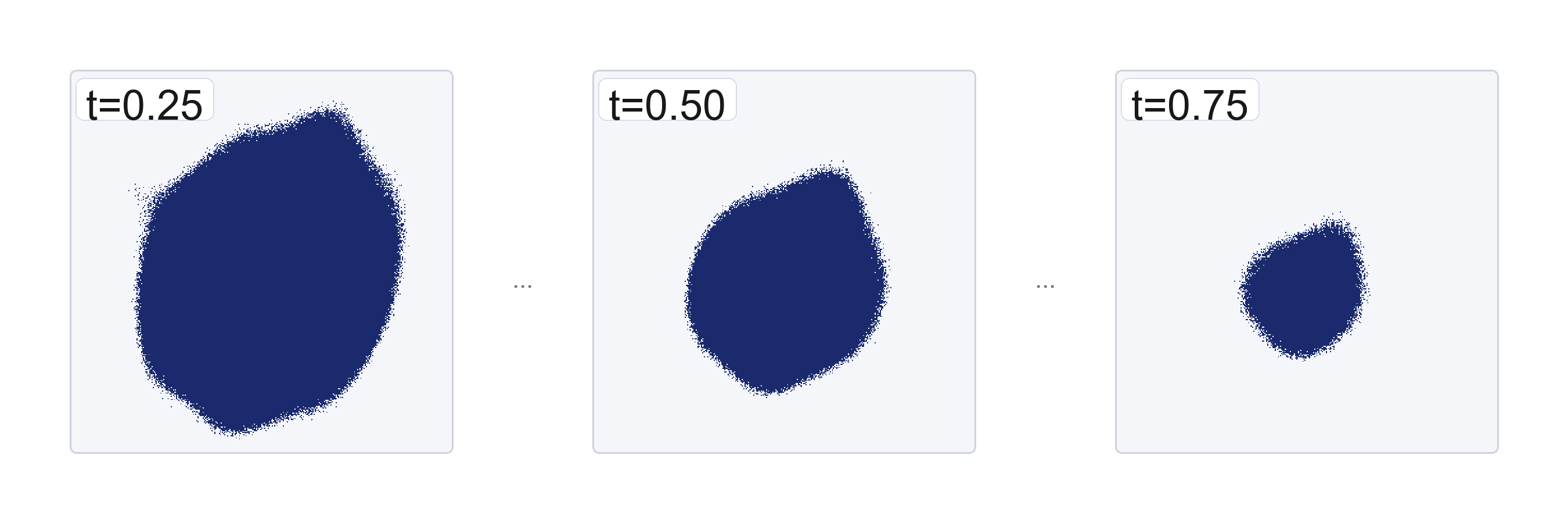}
    \caption{Dynamic}
    \label{fig:thresholding_dynamic}
  \end{subfigure}
  \caption{S1 behaviors generated from the same synthetic Gaussian-like prediction map.
    \subref{fig:thresholding_soft}: The soft setting keeps the gray values $P$.
    \subref{fig:thresholding_fixed} and \subref{fig:thresholding_adaptive}: Fixed and adaptive protocols produce one binarized mask, respectively $B_{1/2}=\mathbf{1}[P>\frac{1}{2}]$ and $B_{t_{\mathrm{adp}}}=\mathbf{1}[P\geq t_{\mathrm{adp}}]$.
    \subref{fig:thresholding_dynamic}: Dynamic protocols evaluate a threshold-indexed sequence of binarized masks $\{B_t=\mathbf{1}[P\geq t]\}_{t\in\mathcal{T}}$.}
  \label{fig:thresholding_behaviors}
\end{figure}

\parhead{Hard Predictions.}
For a binary prediction $B$, \gls{TP} counts target pixels predicted as target, \gls{FP} counts background pixels predicted as target, \gls{TN} counts background pixels predicted as background, and \gls{FN} counts target pixels predicted as background.%
\footnote{Some works formulate background and target as two semantic classes and then report a mean over the two class scores.
Such a two-class mean is not generally equivalent to evaluating only the target channel assumed in \cref{subsec:stage_prediction_treatment}, because the background class contributes its own score and can change the interpretation of the final number.}
For threshold-derived hard partitions, adaptive evaluation counts pixels satisfying $P\geq t_{\mathrm{adp}}$, fixed-binary evaluation commonly counts pixels satisfying $P>\frac{1}{2}$, and dynamic evaluation returns threshold-indexed arrays of \gls{TP}, \gls{FP}, \gls{TN}, and \gls{FN}.
The rule $P>\frac{1}{2}$ is a common fixed-threshold choice and matches the argmax decision of a two-logit output, but it is not an inherent design of fixed-threshold binarization.
A protocol may specify another cutoff or a different boundary convention, such as $P\geq\frac{1}{2}$.
\cref{fig:thresholding_behaviors} illustrates the S1 behaviors of \cref{subsec:stage_prediction_treatment} on one prediction map.

\parhead{Soft Predictions.}
When S1 retains $P(x)\in[0,1]$, each pixel contributes fractional agreement and disagreement according to its prediction value.
With binary \gls{GT} labels $G(x)\in\{0,1\}$, the product-based quantities are
\begin{align}
  \operatorname{TP}_{s} & =\sum_{x} P(x)G(x),        \label{eq:soft_tp}\\
  \operatorname{FP}_{s} & =\sum_{x} P(x)[1-G(x)],    \label{eq:soft_fp}\\
  \operatorname{FN}_{s} & =\sum_{x}[1-P(x)]G(x),     \label{eq:soft_fn}\\
  \operatorname{TN}_{s} & =\sum_{x}[1-P(x)][1-G(x)].
  \label{eq:soft_tn}
\end{align}
The four masses satisfy $N=\operatorname{TP}_{s}+\operatorname{FP}_{s}+\operatorname{FN}_{s}+\operatorname{TN}_{s}=HW$.
They reduce to the corresponding hard counts when $P=B$ is binary.

\subsubsection{Basic Primitive Rates}

Following the standard table-of-confusion logic, the actual target population and actual background population are
\begin{align}
  N_{+} & = \operatorname{TP}+\operatorname{FN} \\
  N_{-} & = \operatorname{TN}+\operatorname{FP}
\end{align}
Four primitive rates normalize the four counts by these two actual populations.
\Gls{TPR} is also called recall, sensitivity, hit rate, probability of detection, or statistical power.
\Gls{FNR} is the miss rate or type-II-error rate.
\Gls{TNR} is specificity or selectivity.
\Gls{FPR} is fall-out, false-alarm rate, probability of false alarm, or type-I-error rate.
The primitive rates are
\begin{align}
  \operatorname{TPR} & = \frac{\operatorname{TP}}{N_{+}} = \frac{\operatorname{TP}}{\operatorname{TP}+\operatorname{FN}} = 1-\operatorname{FNR}
  \\
  \operatorname{FNR} & = \frac{\operatorname{FN}}{N_{+}} = \frac{\operatorname{FN}}{\operatorname{TP}+\operatorname{FN}} = 1-\operatorname{TPR}
  \\
  \operatorname{TNR} & = \frac{\operatorname{TN}}{N_{-}} = \frac{\operatorname{TN}}{\operatorname{TN}+\operatorname{FP}} = 1-\operatorname{FPR}
  \\
  \operatorname{FPR} & = \frac{\operatorname{FP}}{N_{-}} = \frac{\operatorname{FP}}{\operatorname{TN}+\operatorname{FP}} = 1-\operatorname{TNR}
\end{align}
Because these four quantities are class-conditional rates, protocols must specify how to handle empty-target cases where $N_{+}=0$ and empty-background cases where $N_{-}=0$.

\subsubsection{\glsxtrfull{MAE}}

\Gls{MAE} is the most basic and direct expression of pixelwise error.
It keeps the soft prediction map $P$, compares it with the whole-map region target $G$, and uses spatially aligned pixels as the correspondence units.
The resulting absolute-deviation score measures their average discrepancy as
\begin{equation}
  \operatorname{MAE} = \frac{1}{HW}\sum_{x}|P(x)-G(x)|
\end{equation}
Lower \gls{MAE} is better, and the value is reported as raw scalar data.
Its main advantage is that it evaluates the continuous map without requiring threshold selection.
It therefore captures whether the prediction assigns high values to target pixels and low values to background pixels on average.
Its weakness is also inherited from this whole-image averaging.
Large background regions can dilute target errors, and two maps with different structures may obtain similar absolute-error averages if their pixelwise deviations have similar magnitudes.

\parhead{Soft Confusion Statistics.}
Since GT is binary, each absolute error satisfies
\begin{equation}
  |P(x)-G(x)|=P(x)[1-G(x)]+[1-P(x)]G(x).
\end{equation}
Thus, \gls{MAE} can be expressed in terms of the soft confusion quantities as
\begin{equation}
  \operatorname{MAE}
  =\frac{\operatorname{FP}_{s}+\operatorname{FN}_{s}}{N}
  =1-\frac{\operatorname{TP}_{s}+\operatorname{TN}_{s}}{N}.
  \label{eq:soft_mae_confusion}
\end{equation}

\parhead{Size-Invariant Extension.}
\gls{SIMAE}~\cite{SizeInvarianceVariants} preserves the same absolute-error principle but changes the unit over which errors are balanced.
Using the separated-target region branch, it balances absolute errors over encoded target regions and the outside-background region instead of averaging all pixels as one whole image.
If no component satisfies the minimum-area rule of \cref{subsec:stage_entity_extraction}, the largest component or tied largest components are retained.
For each retained component, a target-specific mask $M_k$ is extracted from the bounding-box encoding, giving the local prediction $\tilde P_k=P\odot M_k$ and local \gls{GT} $\tilde G_k=G\land M_k$.
Scores are then computed from these masked maps.
Thus \gls{SIMAE}~\cite{SizeInvarianceVariants} is not a different pixelwise distance, but a target-balanced aggregation of the same absolute differences.

\subsubsection{\glsxtrfull{ROC} Curve}
\glsunset{ROC}

\gls{ROC} curve reporting uses dynamic thresholds to trace paired \gls{FPR} and \gls{TPR} values.
The raw \gls{ROC} curve keeps these threshold-generated points in the \gls{FPR}-\gls{TPR} plane and is therefore a raw-curve reporting mode.
The \gls{ROC} curve and \gls{ROC}-\gls{AUC} use the same upstream evaluation path.
Dynamic thresholds first produce a sequence of hard partitions, each partition yields one \gls{FPR}-\gls{TPR} pair, and these pairs define a curve in rate space.

\gls{ROC}-\gls{AUC} then changes only the S5 report by replacing the raw curve with the area under it.
After the threshold-generated points are ordered along the \gls{FPR} axis, the area summarizes how much target recall can be maintained as the false-positive rate increases.
In the continuous view, this reporting mode is written as
\begin{equation}
  \operatorname{ROC\text{-}AUC} = \int_0^1 \operatorname{TPR}\mathop{}d\operatorname{FPR}
\end{equation}
With finite threshold samples, the integral is usually approximated from the threshold sequence, so \gls{ROC}-\gls{AUC} should be read as a threshold-wise area summary rather than as a new target extraction, matching, or scoring rule.
The meaning of this score is threshold robustness under a target-background rate trade-off.
A better ROC curve stays closer to the upper-left corner of the \gls{FPR}-\gls{TPR} plane, and a larger \gls{ROC}-\gls{AUC} means that target pixels tend to receive higher confidence than background pixels across thresholds.
Therefore, \gls{ROC}-style reporting favors prediction maps with good global ranking and separability between target and background responses, even before a deployment threshold is chosen.

\subsubsection{Precision, Recall, \glsxtrfull{Fbeta}, and \glsxtrfull{AP}}
\glsunset{Fbeta}
\glsunset{AP}

Precision and recall describe complementary sides of target detection and are defined as
\begin{align}
  \operatorname{Precision} & = \frac{\operatorname{TP}}{\operatorname{TP}+\operatorname{FP}}
  \\
  \operatorname{Recall}    & = \operatorname{TPR}
\end{align}
Precision penalizes false target predictions, whereas recall penalizes missed target pixels.
\gls{Fbeta}~\cite{Fmeasure} combines them as
\begin{equation}
  F_{\beta} =
  \frac{(1+\beta^{2})\operatorname{Precision}\operatorname{Recall}}
  {\beta^{2}\operatorname{Precision}+\operatorname{Recall}}
\end{equation}
In much of the \gls{SOD} and \gls{COD} literature~\cite{CMMSODSurvey,Survey-COD}, $\beta^{2}=0.3$~\cite{Fmeasure} is used to emphasize precision, while $\beta^{2}=1$ weights precision and recall equally.
\gls{Fbeta} is an early bridge between pure pixel counting and practical thresholded evaluation.
Common \gls{Fbeta} protocols pair adaptive or dynamic S1 choices with whole-map target extraction, pixelwise correspondence, and an S4 rule that computes \gls{Fbeta} from \gls{TP}, \gls{FP}, \gls{TN}, and \gls{FN}.

\parhead{Dynamic-Threshold Derivatives.}
Under dynamic S1, precision, recall, and \gls{Fbeta} are evaluated over a threshold-indexed sequence.
The resulting \gls{Fbeta} curve and precision-recall curve are S5 raw-curve reports, even when the precision-recall curve is plotted in the precision-recall plane rather than directly against the threshold.
\gls{AP} summarizes the same PR relation by integrating precision over recall:
\begin{equation}
  \begin{aligned}
    \operatorname{AP}
      = \int_{\mathcal{T}}\operatorname{Precision}(t)\mathop{}d\operatorname{Recall}(t)
      = \int_{\mathcal{T}}\operatorname{Precision}(t)
    \frac{d\operatorname{Recall}(t)}{dt}\mathop{}d t
  \end{aligned}
\end{equation}
The first form does not explicitly integrate over thresholds.
However, each PR point is generated by a threshold $t$, so $\operatorname{Precision}$ and $\operatorname{Recall}$ can be treated as threshold-dependent quantities along the ordered threshold path $\mathcal{T}$.
\gls{AP} is thus a scalar summary of a threshold-generated curve obtained by curve integration.
Different protocols may use interpolated precision or numerical integration, but the stage mechanism remains the same.

\parhead{Size-Invariant Extension.}
The size-invariant \gls{Fbeta} variant~\cite{SizeInvarianceVariants} keeps the standard scoring form but changes the evaluation unit before the score is computed.
Like \gls{SIMAE}, it shifts aggregation from whole-image pixel pooling to target-aware local evaluation over retained encoded regions, but the local score remains the standard \gls{Fbeta}.
Although \gls{Fbeta} is not as directly area additive as pixel-error sums, whole-image evaluation can still allow large components to hide failures on small targets during sample-level aggregation.
The size-invariant variant addresses this remaining aggregation bias by computing target-aware local \gls{Fbeta} scores.
Adaptive and binary sample-based variants average retained targets within a sample, whereas dynamic variants preserve per-target threshold curves for downstream summary.

\subsubsection{\glsxtrfull{IoU} and Dice}
\glsunset{IoU}

\gls{IoU}, also known as the Jaccard index, measures target overlap relative to target union:
\begin{equation}
  \operatorname{IoU} = \frac{\operatorname{TP}}{\operatorname{TP}+\operatorname{FP}+\operatorname{FN}}
\end{equation}
Dice is another widely used overlap score closely tied to \gls{IoU}:
\begin{equation}
  \operatorname{Dice} = \frac{2\operatorname{TP}}
  {2\operatorname{TP}+\operatorname{FP}+\operatorname{FN}}
\end{equation}
When computed from the same \gls{TP}, \gls{FP}, and \gls{FN}, Dice can be written as an increasing function of \gls{IoU}:
\begin{equation}
  \operatorname{Dice} = \frac{2\operatorname{IoU}}{1+\operatorname{IoU}}
\end{equation}
Under the same evaluation setting, they produce identical pairwise rankings, although their numerical scales differ.
However, this equivalence holds only when both scores are computed from the same confusion counts.
The nonlinear transform does not generally commute with averaging over images, so separately averaged Dice and \gls{IoU} need not preserve this identity or the same model ranking.
Aggregation after S3 (\cref{subsec:dataset_aggregation}) sums the counts before scoring and therefore preserves both.

\subsubsection{\glsxtrfull{OA}}
\glsunset{OA}

\gls{OA} measures the proportion of correctly predicted pixels among all pixels, making it a direct whole-image average score:
\begin{equation}
  \begin{aligned}
    \operatorname{OA}
      = \frac{\operatorname{TP}+\operatorname{TN}}{\operatorname{TP}+\operatorname{FP}+\operatorname{TN}+\operatorname{FN}}
      = \frac{\operatorname{TP}+\operatorname{TN}}{N_{+}+N_{-}}
      = 1-\frac{\operatorname{FP}+\operatorname{FN}}{N_{+}+N_{-}}
  \end{aligned}
\end{equation}
The last form exposes its link to \gls{MAE}.
For binary predictions and binary \gls{GT}, $\operatorname{OA}=1-\operatorname{MAE}$.
For soft predictions, $1-\operatorname{MAE}$ gives the corresponding soft whole-image correctness score.
This shared whole-image averaging is also the source of its weakness in target segmentation.
\gls{OA} and its soft analogue emphasize the prediction correctness of all pixels, whereas target segmentation evaluation often cares more about the target region itself.
When target pixels occupy only a small area, a prediction can obtain a high whole-image correctness score by matching most background pixels while still missing or distorting the target, so this score can be misleading and is usually reported less often than target-sensitive alternatives.

\subsubsection{\glsxtrfull{BER}}
\glsunset{BER}

\gls{BER}~\cite{TransparentObjectSegmentation,SurveyShadowDetection} is a balance-oriented confusion-matrix metric used in \gls{TOS}, \gls{SD}, and related binary target segmentation settings where target and background regions can be highly imbalanced.
In the common protocol, it starts from a fixed-threshold binary mask and compares it with the whole-map region target through pixelwise correspondence.
\gls{OA} weights each pixel equally, so a large background region can dominate the final score.
\gls{BER} instead gives the target class and the background class equal class-level weight by measuring the error rate of each side separately and then averaging the two errors:
\begin{equation}
  \begin{aligned}
    \operatorname{BER}
      = 1-\frac{1}{2}\left(\operatorname{TPR}+\operatorname{TNR}\right)
      = \frac{1}{2}\left(\operatorname{FNR}+\operatorname{FPR}\right)
  \end{aligned}
\end{equation}
Lower values are better, with zero meaning that both target pixels and background pixels are classified perfectly.
A fixed operating point produces a single scalar value, which is reported as raw scalar data.
A method cannot obtain a good \gls{BER} score merely by predicting most pixels as background, because that behavior keeps \gls{FPR} low but makes \gls{FNR} high.
Conversely, over-segmenting the target region may reduce missed target pixels but increases false alarms in the background, which raises the \gls{FPR} term.
The metric favors simultaneous target recovery and background rejection.
\gls{BER} remains a pixel-counting metric.
It corrects the class-prior weighting of accuracy, but it does not evaluate object topology, boundary fidelity, connected-component matching, or the spatial layout of errors.

\subsubsection{\glsxtrfull{Em}}
\glsunset{Em}

\gls{Em}~\cite{Emeasure} was proposed for binary foreground-map evaluation.
It combines pixelwise agreement with the global foreground proportions of the prediction and \gls{GT}.
For a soft prediction map, adaptive thresholding yields one binary mask and one score.
A threshold sweep produces a score curve, which can be retained or summarized by its threshold-wise mean or maximum.

\parhead{Enhanced Alignment.}
Let $N=HW$, $B=B_t\in\{0,1\}^{H\times W}$, and $G\in\{0,1\}^{H\times W}$.
Write $p=\bar B$ and $g=\bar G$ for their foreground proportions.
The method keeps spatially aligned pixel correspondence and centers both masks by their whole-map means.
For $0<g<1$, the pixelwise alignment and enhanced score are
\begin{align}
  a(x)     & =\frac{2[B(x)-p][G(x)-g]}{[B(x)-p]^2+[G(x)-g]^2}, \\
  E_m(B,G) & =\frac1N\sum_{x}\frac{[1+a(x)]^2}{4}.
  \label{eq:em_pixel_definition}
\end{align}
A positive $a(x)$ means that $B(x)$ and $G(x)$ deviate from their own means in the same direction.
A negative value means that they disagree after centering.
The quadratic enhancement maps alignment values from $[-1,1]$ to $[0,1]$.
The mapping reduces the effect of negative alignment and strengthens positive alignment before spatial averaging.
Higher $E_m$ indicates better agreement.
Besides, all-background and all-foreground \gls{GT} masks bypass the alignment term.
Under the same mean normalization, their scores are defined separately as $\operatorname{TN}/N=1-p$ and $\operatorname{TP}/N=p$, respectively.
The finite-image normalization divides these sums by $N-1$ instead of $N$, so a perfect prediction can score slightly above one.
The main strength of \gls{Em} is that it keeps pixelwise comparison simple while adding global centering.
The original study reported reduced preference for random noise and generic foreground maps~\cite{Emeasure}.
The score still depends on binarization and uses only a whole-map mean as context.
It does not explicitly model target instances, topology, or boundary tolerance.

\parhead{Connection to Confusion Statistics.}
For binary images, the foreground proportions can be expressed through the confusion counts as $p = \frac{\operatorname{TP}+\operatorname{FP}}{N}$ and $g = \frac{\operatorname{TP}+\operatorname{FN}}{N}$.
Let $(\zeta,\eta)\in\{0,1\}^2$ denote the prediction and \gls{GT} values at the same pixel.
Their alignment and enhanced response are
\begin{align}
  A_{\zeta\eta}(p,g)    & =\frac{2(\zeta-p)(\eta-g)}{(\zeta-p)^2+(\eta-g)^2}, \\
  \Phi_{\zeta\eta}(p,g) & =\frac{[1+A_{\zeta\eta}(p,g)]^2}{4}.
\end{align}
The pairs $(1,1)$, $(1,0)$, $(0,1)$, and $(0,0)$ correspond to \gls{TP}, \gls{FP}, \gls{FN}, and \gls{TN}, respectively.
Within each class, every pixel has the same alignment $A_{\zeta\eta}$ and enhanced response $\Phi_{\zeta\eta}$.
Substituting $p$ and $g$ gives the four alignment values in terms of confusion counts
\begin{align}
  A_{11} & =\frac{2(1-p)(1-g)}{(1-p)^2+(1-g)^2} = \frac{2(\operatorname{FN}+\operatorname{TN})(\operatorname{FP}+\operatorname{TN})}{(\operatorname{FN}+\operatorname{TN})^2+(\operatorname{FP}+\operatorname{TN})^2}, \\
  A_{10} & =\frac{-2(1-p)g}{(1-p)^2+g^2} = \frac{-2(\operatorname{FN}+\operatorname{TN})(\operatorname{TP}+\operatorname{FN})}{(\operatorname{FN}+\operatorname{TN})^2+(\operatorname{TP}+\operatorname{FN})^2}, \\
  A_{01} & =\frac{-2p(1-g)}{p^2+(1-g)^2} = \frac{-2(\operatorname{TP}+\operatorname{FP})(\operatorname{FP}+\operatorname{TN})}{(\operatorname{TP}+\operatorname{FP})^2+(\operatorname{FP}+\operatorname{TN})^2}, \\
  A_{00} & =\frac{2pg}{p^2+g^2} = \frac{2(\operatorname{TP}+\operatorname{FP})(\operatorname{TP}+\operatorname{FN})}{(\operatorname{TP}+\operatorname{FP})^2+(\operatorname{TP}+\operatorname{FN})^2}.
\end{align}
The enhanced pixel response is $\Phi_{B(x)G(x)}(p,g)$.
For example, the $\operatorname{TP}$ pixels with $(B(x),G(x))=(1,1)$ all contribute $\Phi_{11}$, giving a total of $\operatorname{TP}\Phi_{11}$.
Applying the same grouping to all four classes gives
\begin{equation}
  \begin{aligned}
    E_m(B,G)
     & =\frac{1}{N}\sum_{x}\Phi_{B(x)G(x)}(p,g)       \\
     & =\frac{1}{N}\Biggl(
          \sum_{\substack{B(x)=1\\G(x)=1}}\Phi_{11}
          +\sum_{\substack{B(x)=1\\G(x)=0}}\Phi_{10}
          +\sum_{\substack{B(x)=0\\G(x)=1}}\Phi_{01}
          +\sum_{\substack{B(x)=0\\G(x)=0}}\Phi_{00}
          \Biggr)                                         \\
     & =\frac{
          \operatorname{TP}\Phi_{11}
          +\operatorname{FP}\Phi_{10}
          +\operatorname{FN}\Phi_{01}
          +\operatorname{TN}\Phi_{00}}{N}.
  \end{aligned}
  \label{eq:em_confusion_reduction}
\end{equation}
Thus $E_m$ is a nonlinear function of the four confusion counts.
Binary prediction and \gls{GT} pairs with the same confusion counts therefore receive the same score, even when their spatial layouts differ.

\subsubsection{\glsxtrfull{wFbeta}}
\glsunset{wFbeta}

\gls{wFbeta}~\cite{wFmeasure} extends classical \gls{Fbeta} by accounting for spatial dependencies and pixel importance.
It compares the soft prediction $P$ with the whole-map binary \gls{GT} $G$ using pixelwise correspondence.
Conventional \gls{AUC}, \gls{AP}, \gls{Fbeta}, and \gls{IoU} can produce counterintuitive rankings for target maps, a limitation originally discussed for foreground maps~\cite{wFmeasure}.
\gls{wFbeta} starts from the observation that pixel errors do not always have the same perceptual importance.
Errors near target boundaries, errors deep inside target regions, and scattered false positives in background should not necessarily be treated as independent unit mistakes.

\parhead{Contextual Aggregation.}
Isolated foreground errors can receive smaller penalties when neighboring foreground pixels are correctly predicted.
To account for this local dependence, the metric builds an auxiliary map from $E=|P-G|$ by replacing each background error with the error at the nearest \gls{GT} foreground pixel.
It smooths this map with a Gaussian kernel and retains the smaller of the original and smoothed errors at foreground pixels.
Background pixels keep their original errors, giving the effective error map $\widetilde E$.

\parhead{Contribution Weighting.}
False positives farther from the target receive larger penalties.
The metric therefore defines the weights as
\begin{equation}
  W(x)=
  \begin{cases}
    1,                                              & G(x)=1, \\
    2-\exp\!\left[\dfrac{\ln(0.5)}{5}d(x)\right], & G(x)=0,
  \end{cases}
  \label{eq:weighted_f_weights}
\end{equation}
where $d(x)$ is the Euclidean distance in pixels from $x$ to the nearest \gls{GT} foreground pixel.
The weighted error map is $E^w=\widetilde E\odot W$.
The weighted confusion quantities are defined as
\begin{align}
  \operatorname{TP}^{w} & = \sum (1-E^{w})G     \\
  \operatorname{TN}^{w} & = \sum (1-E^{w})(1-G) \\
  \operatorname{FP}^{w} & = \sum E^{w}(1-G)     \\
  \operatorname{FN}^{w} & = \sum E^{w}G
\end{align}
Weighted precision and weighted recall share the same weighted true-positive term and are combined in the same \gls{Fbeta} form as the classical protocol, producing \gls{wFbeta} as raw scalar data.
In \gls{SOD}, \gls{wFbeta} uses $\beta^{2}=1$, whereas classical \gls{Fbeta} commonly uses $\beta^{2}=0.3$.

\subsubsection{\glsxtrfull{MSIoU}}
\glsunset{MSIoU}

Standard \gls{IoU} can underrepresent thin structures and fine boundaries.
\gls{MSIoU}~\cite{MSIoU} compares detected and \gls{GT} regions at multiple pyramid levels to increase sensitivity to fine structures.
In the protocol summarized here, it starts from a fixed binary detected region, but changes the evaluated target representation from dense regions to whole-map edge maps.
The metric extracts these edge maps using Sobel filters, then applies the pyramid modifier through grid-based shrinking at multiple levels and integration of the resulting base scores.
At each level, it keeps spatially aligned pixelwise correspondence and computes multiscale target-edge coverage by reusing \gls{TPR}.

At level $j$, let $E_B^{(j)}$ and $E_G^{(j)}$ denote the prediction and \gls{GT} edge maps shrunk by cells of side $2^j$.
The level-wise score reuses the primitive \gls{TPR} from confusion-matrix metrics:
\begin{equation}
  r_j = \operatorname{TPR}^{(j)} = \frac{|E_B^{(j)}\cap E_G^{(j)}|}{|E_G^{(j)}|}
\end{equation}
Although the metric name contains \gls{IoU}, the denominator of $r_j$ counts only \gls{GT} edge pixels instead of the union of both edge maps, so $r_j$ takes the \gls{TPR} form and measures target-edge coverage.
Both counts are incremented by one during computation, so an empty \gls{GT} edge map gives $r_j=1$.
With $J$ levels, $j=0,\ldots,J-1$, as in \cref{sec:metric_construction}, let $r(u)$ linearly interpolate the samples $(u_j,r_j)$ at the normalized scale coordinate $u_j=j/(J-1)$.
\gls{MSIoU} integrates this level-indexed \gls{TPR}:
\begin{equation}
  \operatorname{MSIoU} = \int_{0}^{1} r(u)\mathop{}d u
\end{equation}
Because $r(u)$ is piecewise linear, the integral equals the trapezoidal sum over the $J$ levels and yields one raw scalar value.
For $J=1$, where $u_0$ is undefined, \gls{MSIoU} is directly defined as $r_0$.
\gls{MSIoU} inherits the interpretability of overlap metrics while becoming more sensitive to fine structures and boundary fragments.

\subsubsection{\glsxtrfull{hIoU}}
\glsunset{hIoU}

\gls{hIoU}~\cite{HierarchicalIoU} was proposed for \gls{IRSTD}~\cite{IRSTDSurvey}, a setting where targets are often extremely small, sparsely distributed, and easily confused with clutter or bright background interference.
These task characteristics make target localization an important part of performance, in addition to pixel-level segmentation quality.
They motivate a metric that accounts explicitly for matched, missed, and false-alarm targets.
The metric modifies the whole-map \gls{IoU} protocol conventionally used in \gls{IRSTD} by first parsing connected target components from the thresholded prediction and the \gls{GT}.
Its target matching rule then assigns prediction components to \gls{GT} components by combining overlap qualification, Hungarian assignment with centroid distance as the cost, and distance-based compensation for otherwise unmatched small targets.
With this design, one-to-one target correspondence becomes an explicit part of the metric.

Let $\mathcal{M}$ denote the matched target pairs, and let $\mathrm{TP}_{tgt}$, $\mathrm{FP}_{tgt}$, and $\mathrm{FN}_{tgt}$ denote the numbers of matched, unmatched predicted, and unmatched \gls{GT} objects.
For each matched pair $\pi\in\mathcal{M}$ of a \gls{GT} component and a predicted component, let $q_{\pi}$ denote their \gls{IoU}, as in \cref{tab:stage_metric_formulas}.
The usual discrete form can be read as a target-localization factor multiplied by the mean segmentation quality of matched pairs:
\begin{equation}
  \begin{aligned}
    \operatorname{hIoU}
     =
    \frac{\mathrm{TP}_{tgt}}{\mathrm{TP}_{tgt}+\mathrm{FP}_{tgt}+\mathrm{FN}_{tgt}}
    \cdot
    \frac{\sum_{\pi\in\mathcal{M}} q_{\pi}}{\mathrm{TP}_{tgt}}
     =
    \frac{\sum_{\pi\in\mathcal{M}} q_{\pi}}
    {\mathrm{TP}_{tgt}+\mathrm{FP}_{tgt}+\mathrm{FN}_{tgt}}
  \end{aligned}
\end{equation}
When no target pair is matched, the segmentation term is treated as zero.
Under the common \gls{hIoU} protocol, a fixed threshold gives a single \gls{hIoU} value that is reported as raw scalar data.
\gls{hIoU} adds separated target extraction and explicit target matching before overlap scores are formed, while the pairwise overlap score remains \gls{IoU}.

\parhead{Integral View.}
The same \gls{hIoU} score can be rewritten as an integral over post-matching overlap quality.
This view explains why the product between target-localization quality and matched-pair segmentation quality is more than a heuristic multiplication.
Once the original matching protocol fixes $\mathcal{M}$, the topological capacity $D=\mathrm{TP}_{tgt}+\mathrm{FP}_{tgt}+\mathrm{FN}_{tgt}$ is frozen as a constant, and the auxiliary variable $\tau$ only probes the overlap values $q_{\pi}$ of the fixed pairs.
This static-topology probing separates topology resolution from quality profiling.
Missed and false-alarm targets set the baseline penalty through $D$, while matched-pair quality is accumulated over all post-matching quality levels.
Dynamic-threshold protocols would recompute candidate edges or assignments as the threshold changes.
Here, the graph is fixed after the original matching step and is not re-matched inside the integral.
Starting from the discrete segmentation term, each $q_{\pi}\in[0,1]$ admits the representation
\begin{equation}
  q_{\pi} = \int_0^1 \mathbf{1}[\tau<q_{\pi}]\mathop{}d\tau
\end{equation}
Define the static-topology survival count
\begin{equation}
  \begin{aligned}
      T_{\mathcal{M}}(\tau)
      = \sum_{\pi\in\mathcal{M}}\mathbf{1}[\tau<q_{\pi}]
      = |\{\pi\in\mathcal{M}\mid \tau<q_{\pi}\}|
  \end{aligned}
\end{equation}
as the number of resolved matched pairs whose overlap quality remains above level $\tau$.
Substituting the integral representation of $q_{\pi}$ into the discrete \gls{hIoU} formula and exchanging the finite sum with the integral gives
\begin{equation}
  \begin{aligned}
    \operatorname{hIoU}
     = \frac{1}{D}\sum_{\pi\in\mathcal{M}} q_{\pi}
     = \frac{1}{D}\sum_{\pi\in\mathcal{M}}\int_0^1 \mathbf{1}[\tau<q_{\pi}]\mathop{}d\tau
     = \frac{1}{D}\int_0^1 T_{\mathcal{M}}(\tau)\mathop{}d\tau
     = \int_0^1 \mathcal{S}_{\mathcal{M}}(\tau)\mathop{}d\tau
  \end{aligned}
\end{equation}
where $\mathcal{S}_{\mathcal{M}}(\tau)=T_{\mathcal{M}}(\tau)/D$ is a static-topology survival score, not a dynamic \gls{IoU} curve.
The integral form gives a theoretical reading of the original discrete \gls{hIoU}.
Localization errors fix the denominator once, and segmentation quality contributes through continuous quality retention over the fixed matched topology.

\subsection{\glsxtrfull{Sm}}
\glsunset{Sm}

\gls{Sm}~\cite{Smeasure} was proposed to evaluate target maps through object-aware and region-aware structural similarity.
It keeps the soft prediction map and the whole-map region target, with spatially aligned pixels as the correspondence units.
Its structural scoring rule combines an object-aware term and a region-aware term:
\begin{equation}
  S_m = \alpha S_o+(1-\alpha)S_r
\end{equation}
where $S_o$ measures object-level similarity and $S_r$ measures region-level similarity.
A common default is $\alpha=\frac{1}{2}$.
When $G$ contains no target or no background pixels, $S_m$ reduces to $1-\bar P$ or $\bar P$, respectively.
The resulting $S_m$ is reported as raw scalar data.

\parhead{Object-Aware Score.}
$S_o$ compares prediction statistics separately over target and background regions.
It uses the mean and standard deviation of $P$ over target pixels in $G$ for foreground comparison, and those of $1-P$ over background pixels in $G$ for background comparison.
The object-level foreground and background similarities can be written as
\begin{align}
  O_{FG} & = \frac{2\mu(P\mid G=1)}{\mu(P\mid G=1)^2+1+2\lambda\sigma(P\mid G=1)}
  \\
  O_{BG} & = \frac{2\mu(1-P\mid G=0)}{\mu(1-P\mid G=0)^2+1+2\lambda\sigma(1-P\mid G=0)}
\end{align}
where $\mu(\cdot)$ and $\sigma(\cdot)$ denote the mean and standard deviation over the indicated pixels, and $\lambda=\frac{1}{2}$ balances the mean-response and variance terms.
If $\rho$ is the target-pixel ratio in $G$, the final object-aware score is
\begin{equation}
  S_o = \rho O_{FG}+(1-\rho)O_{BG}
\end{equation}
Through these two terms, $S_o$ rewards high and consistent responses inside the annotated target while also accounting for background suppression.

\parhead{Region-Aware Score.}
$S_r$ measures object-part structure by partitioning both the prediction map and the \gls{GT} map with horizontal and vertical cut lines that intersect at the centroid of the \gls{GT} foreground.
This partition yields four blocks.
For each block $k$, a \gls{SSIM}-like~\cite{SSIM} similarity $\operatorname{ssim}(k)$ is computed independently.
Let $\Omega_k$ denote block $k$ and set its weight to the block-area ratio $w_k=|\Omega_k|/(HW)$, so that $\sum_{k=1}^{4}w_k=1$.
The region-aware score is
\begin{equation}
  S_r = \sum_{k=1}^{4}w_k\operatorname{ssim}(k)
\end{equation}
The weighted partitioning emphasizes whether the prediction preserves the spatial layout and object-part structure of the target.

Overlap and average-error scores do not necessarily reveal whether a prediction preserves the target as a coherent structure.
\gls{Sm}~\cite{Smeasure} keeps soft whole-map comparison and replaces pure pixelwise error or overlap with object-aware foreground-background statistics and region-aware layout similarity.
It therefore rewards predictions that preserve coherent target structure.

\subsection{\glsxtrfull{HCE}}
\glsunset{HCE}

\gls{HCE}~\cite{HumanCorrectionEffortMeasure} estimates the amount of manual work needed to correct a predicted mask.
It changes the evaluation question.
Instead of asking how many pixels are wrong, it asks how difficult the remaining errors may be for a human editor.
Under the fixed-threshold \gls{HCE} protocol, the prediction is first binarized with the rule $P>\frac{1}{2}$.
The metric then computes one image-level effort value for the resulting binary mask.
False-positive and false-negative regions are relaxed morphologically to avoid overcounting minor boundary disagreements.
Whole-map \gls{GT} skeletonization preserves structurally important missed components, so the target representation is treated as a whole-map skeleton setting, with pixelwise correspondence retained during error extraction.
After this relaxation, \gls{HCE} separates the remaining errors into residual \gls{FP} regions that should be removed and residual \gls{FN} regions that should be recovered.
For each error type, it counts two forms of correction work.
The subscript $\mathrm{poly}$ counts polygon control points obtained from the approximated correction contours.
The subscript $\mathrm{ind}$ counts independent connected error regions that require separate region-level correction.
The superscripts $\mathrm{FP}$ and $\mathrm{FN}$ indicate whether the counted operations belong to extra predicted regions or missed target regions.
The edit-cost scoring rule sums these four correction counts with unit operation costs:
\begin{equation}
  \operatorname{HCE}
  = N_{\mathrm{poly}}^{\mathrm{FP}}+N_{\mathrm{ind}}^{\mathrm{FP}}
  + N_{\mathrm{poly}}^{\mathrm{FN}}+N_{\mathrm{ind}}^{\mathrm{FN}}
\end{equation}
$N_{\mathrm{poly}}^{\mathrm{FP}}$ and $N_{\mathrm{ind}}^{\mathrm{FP}}$ estimate deletion effort for false-alarm structures, while $N_{\mathrm{poly}}^{\mathrm{FN}}$ and $N_{\mathrm{ind}}^{\mathrm{FN}}$ estimate recovery effort for missed structures.
A lower \gls{HCE} value is better.
The resulting image-level effort is reported as raw scalar data, and zero means that no residual correction operation is counted by this protocol.
\gls{HCE} complements overlap measures by estimating an editing cost that overlap scores cannot express.
Two masks with similar pixel overlap can require different correction effort if one contains many disconnected mistakes and the other contains one simple boundary shift.
The score depends on this modeling of the editing cost, so its interpretation is tied to the correction operations that it counts.

\subsection{\glsxtrfull{FT}}
\label{sup:ft}
\glsunset{FT}

\gls{FT}~\cite{FractionalThreshold} measures the fraction of predictions meeting a quality threshold calibrated from human segmentation agreement.
For a higher-is-better base score $q_i$ of sample $i$, the single-sample acceptance indicator is
\begin{equation}
  a_i = \mathbf{1}[q_i\geq q_{\mathrm{hum}}]
  \label{eq:ft_acceptance}
\end{equation}
where $q_{\mathrm{hum}}$ is the human reference threshold.
The original evaluation~\cite{FractionalThreshold} sets this threshold to the first percentile of human-to-human score distributions and applies the criterion to Dice.
This adds human calibration to an existing scoring rule and accommodates different base metrics.
For $N_{\mathrm{img}}$ samples, the dataset-level acceptance fraction is
\begin{equation}
  \operatorname{\glsentryshort{FT}}(q_{\mathrm{hum}})
  = \frac{1}{N_{\mathrm{img}}}\sum_{i=1}^{N_{\mathrm{img}}}a_i
  \label{eq:ft_dataset}
\end{equation}
The formula in \cref{tab:stage_metric_formulas} gives the single-sample indicator in \cref{eq:ft_acceptance}, while \cref{eq:ft_dataset} averages these indicators across samples.
The same predictions can yield different acceptance fractions under different reference calibrations.
Reproducibility therefore requires specifying the base metric, human reference cohort, and threshold-calibration rule.

\subsection{\glsxtrfull{Cm}}
\glsunset{Cm}

The \gls{COD} setting~\cite{Survey-COD,ConcealedObjectDetection,COD10K} introduces a difficulty not captured by ordinary target segmentation metrics.
An object can be hard because of its relation to the surrounding background.
\gls{Cm}~\cite{ContextMeasure} was proposed to make this contextual dependency explicit.
It is a soft metric combining forward inference and reverse deduction.
Following the original notation but reusing $P$ and $G$, the forward map $F(G\mid P)$ measures how the prediction activates \gls{GT}-induced contextual relevance, and the reverse map $R(P\mid G)$ measures how completely the \gls{GT} can be inferred back from the prediction.
With a Gaussian context kernel $\mathcal{K}$ estimated from the \gls{GT} target distribution, the efficient form uses the approximations
\begin{align}
  F(G\mid P) & \approx P\odot(\mathcal{K}\ast G)                                                      \\
  R(P\mid G) & \approx \frac{e}{e-1}G\odot\left[\mathbf{1}-\exp\left(-\mathcal{K}\ast P\right)\right]
\end{align}
The image-level forward and reverse scores are then normalized as
\begin{align}
  F_m & = \frac{\|F(G\mid P)\|_1}{\|P\|_1} \\
  R_m & = \frac{\|R(P\mid G)\|_1}{\|G\|_1}
\end{align}
and combined into the concrete \gls{Cm} score in a \gls{Fbeta}-like form
\begin{equation}
  C_{\beta} = \frac{(1+\beta^{2})F_mR_m}{\beta^{2}F_m+R_m}
\end{equation}
\gls{Cm} keeps a soft prediction map, a whole-map region target, and spatially aligned pixels, but changes S4 into a context-dependent relevance calculation.
It extends evaluation beyond pixel discrepancy, overlap, and structure toward task-conditioned scoring, where the quality of a \gls{COD} prediction depends on how its responses relate to the spatial context of the annotated object.

\parhead{Bidirectional Response.}
The response totals $\|F(G\mid P)\|_1$ and $\|R(P\mid G)\|_1$ are generally different.
With the denominators $\|P\|_1$ and $\|G\|_1$ specified above, they therefore do not act as a shared true-positive mass for precision and recall from one confusion matrix.
The combination operates on the separately constructed directional responses, giving the bidirectional response base mode.

\parhead{Camouflage-Aware Weighting.}
The weighted extension further incorporates the original RGB image because camouflage is not determined by the mask alone.
It estimates a camouflage degree map $\Gamma$ by matching target patches to surrounding background patches in Lab color space with spatial constraints.
Larger values in $\Gamma$ indicate that a target region is visually closer to its context, so correct recovery in these regions should receive stronger credit than recovery in visually obvious regions.
The weight is applied only to the reverse term because the added cue is intended to measure how well the annotated target can be recovered under image-conditioned camouflage difficulty.
The forward term still measures prediction relevance under the context kernel and is kept unchanged to avoid giving extra credit to context-like false responses outside the target.
Specifically, the weighted reverse score $R_w$ is defined as
\begin{equation}
  R_w = \frac{\|R(P\mid G)\odot(G+\Gamma)\|_1}{\|G+\Gamma\|_1}
  \label{eq:context_weighted_reverse}
\end{equation}
Replacing $R_m$ with $R_w$ in $C_{\beta}$ yields the camouflage-aware \gls{wCm}.
The default settings~\cite{ContextMeasure} use $\beta^2=1$ for $C_{\beta}$ and $\beta^2=1.2$ for \gls{wCm}.
This design keeps the forward term and the \gls{Fbeta}-like combination unchanged while making the final score depend on both mask agreement and scene context.

\parhead{Contextual Aggregation \& Contribution Weighting.}
The forward term can be read as prediction responses weighted by $\mathcal K\ast G$, while the reverse term aggregates neighboring prediction responses through $\mathcal K\ast P$ and applies nonlinear saturation.
The convolutions aggregate contextual evidence to construct the forward and reverse responses combined by the bidirectional response score.
The camouflage map $\Gamma$ supplies contribution weights for the reverse response.
Both the base response and this coefficient use context, but they play different roles in the score construction.

\subsection{Surface-Distance Metrics}
\label{sup:surface_distance}

Representative metrics, \ie, \gls{ASD}, \gls{ASSD}, \gls{HD}, and \gls{HD95}, measure boundary displacement using nearest-neighbor distances.
The definitions below give each boundary point equal weight.
All these metrics report a scalar distance, with smaller values indicating closer boundary agreement.
These metrics share spatial-distance statistics as the base mode, with the formulas in \cref{tab:stage_metric_formulas} specifying the particular summary.
The shared construction exposes boundary displacement, with the mean emphasizing overall alignment and the maximum or percentile emphasizing larger local errors.

\parhead{Boundary Extraction \& Matching.}
Let $\partial B$ and $\partial G$ be the contour point sets of a binary prediction $B$ and binary \gls{GT} $G$ in the image plane.
Points from all components of the evaluated class are compared together.
Each $x\in\partial B$ is matched to its nearest point in $\partial G$, at distance $\min_{y\in\partial G}\|x-y\|_2$.
Exchanging the boundaries gives the reverse direction.
Multiple points may share a match, and each source point contributes one distance.

\subsubsection{\glsxtrfull{ASD}}
\label{sup:asd}

\gls{ASD} measures the mean distance from a predicted boundary point to its nearest reference point.
\begin{equation}
  \operatorname{\glsentryshort{ASD}}(B,G)
  =\frac{1}{|\partial B|}\sum_{x\in\partial B}\min_{y\in\partial G}\|x-y\|_2.
  \label{eq:asd}
\end{equation}
The score is generally asymmetric and can remain small when part of the reference boundary is missing.

\parhead{Symmetric Extension.}
The symmetric extension \gls{ASSD} averages over points on both boundaries.
\begin{equation}
  \operatorname{\glsentryshort{ASSD}}(B,G)
  =\frac{\displaystyle\sum_{x\in\partial B}\min_{y\in\partial G}\|x-y\|_2
    +\displaystyle\sum_{y\in\partial G}\min_{x\in\partial B}\|y-x\|_2}
  {|\partial B|+|\partial G|}.
  \label{eq:assd}
\end{equation}
The two scores can also be averaged as $\frac{1}{2}[\operatorname{\glsentryshort{ASD}}(B,G)+\operatorname{\glsentryshort{ASD}}(G,B)]$.
This gives each direction equal weight and can differ from \cref{eq:assd} when $|\partial B|\ne|\partial G|$.

\subsubsection{\glsxtrfull{HD}}
\label{sup:hd}

\gls{HD} measures the largest boundary displacement across both directions.
\begin{equation}
  \operatorname{\glsentryshort{HD}}(B,G)
  =\max\left\{
  \max_{x\in\partial B}\min_{y\in\partial G}\|x-y\|_2,
  \max_{y\in\partial G}\min_{x\in\partial B}\|y-x\|_2
  \right\}.
  \label{eq:hd}
\end{equation}
An isolated distant point or component can therefore determine the score.

\parhead{Percentile Extension.}
\gls{HD95} takes the 95th percentile in each direction and returns the larger value.
\begin{equation}
  \operatorname{\glsentryshort{HD95}}(B,G)
  =\max\left\{
    \underset{x\in\partial B}{\operatorname{percentile}_{95}}\!\left(\min_{y\in\partial G}\|x-y\|_2\right),
    \underset{y\in\partial G}{\operatorname{percentile}_{95}}\!\left(\min_{x\in\partial B}\|y-x\|_2\right)
  \right\}.
  \label{eq:hd95}
\end{equation}
Each percentile is computed over distances from the source boundary points.
For $n$ sorted values, we use linear interpolation at rank $1+0.95(n-1)$, with ranks starting at one.
Replacing maxima with percentiles reduces sensitivity to isolated extreme distances.
An alternative takes a single 95th percentile over the distances from both boundaries together, which dilutes errors confined to one direction (\cref{tab:surface_protocols}).

\subsection{\glsxtrfull{SegAE}}
\label{sup:segae}
\glsunset{SegAE}

\gls{SegAE}~\cite{SegAE} predicts mask quality as $\hat q=\mathcal{Q}_{\theta}(I,B,c)$, where $I$ is the CT image, $B$ is the binary candidate prediction, and $c$ identifies the anatomical class.
The original method uses the supervision target $q=\operatorname{Dice}(B,G)$, with $G$ denoting the \gls{GT} mask, and combines a regression loss with an optimal-pair ranking loss during training.
At inference, the evaluator estimates this target score from $I$, $B$, and $c$ without access to $G$.
The evaluator jointly encodes the image and mask and uses class text embeddings to condition quality regression.
The prediction for the specified class is evaluated as one region target.
Within our framework, we interpret the class-conditioned assessment as implicit learned matching and the quality regression as implicit learned scoring.
These roles are implemented within the same evaluator and it reports the resulting quality estimate as a scalar.
The learned-evaluation formulation can in principle use another quality target, but this would require corresponding supervision and training rather than changing the interpretation of the existing model output.

\newpage

\section{Experiment Details and Additional Analyses}
\label{sup:controlled_setup}

\subsection{Shared Benchmark}
\label{sup:protocol_benchmark}

The shared benchmark in \cref{subsec:protocol_sensitivity} contains 300 \gls{GT} masks, 60 in each of five groups, \ie, empty maps, small targets, large targets, multiple targets, and thin structures.
Each mask has four prediction types, giving the 1,200 prediction-\gls{GT} pairs scored in \cref{tab:controlled_cases}.

\subsubsection{GT Mask Sampling Details}
In each group, 48 masks are ordinary and 12 are additional.
The additional masks of the four nonempty groups carry the cutoff-equality cases in \cref{sup:prediction_synthesis}.
Empty-group masks contain no target pixels.
Small and large targets are sampled from disks, ellipses, rotated rectangles, and lobed shapes with equal probability.
Small-target and large-target scales are drawn uniformly from $[3,11]$ and $[20,43]$ pixels, respectively.
The scale is the disk radius, ellipse major semiaxis, rectangle half-length, or major semiaxis of the underlying lobed shape.
Noncircular targets use aspect ratios in $[0.5,1]$ and orientations in $[0,2\pi)$.
Lobed boundaries modulate the normalized radius by $1+a\cos(k\theta)$, with $a\in[0.10,0.25]$ and $k\in\{3,4,5\}$.
Rasterization retains the largest four-connected component of each shape.
Multiple-target masks contain two to six components, with one scale in $[14,26]$, one in $[3,7]$, and any remaining scales in $[5,17]$ pixels.
We sample their centers within the image and reject placements that violate a sampled clearance of 2--9 pixels from existing components.
Thin structures are straight, curved, or branched with equal probability.
The main paths of thin structures have lengths of 45--115 pixels before bending, random orientations, and dilation radii of zero to two pixels.
Curved and branched paths use a quadratic bend of 8--25 pixels, and branched structures add one to three branches of varying length and position.
All paths are rasterized with four-connected steps and placed with a margin from the image boundary.

\subsubsection{Prediction Synthesis Details}
\label{sup:prediction_synthesis}

\parhead{Confidence, Smoothing, \& Noise.}
Each prediction type smooths its constructed binary mask with a sampled Gaussian scale and maps zero and one to sampled background and foreground levels.
\ptype{Balanced Response} uses foreground levels in $[0.8,1]$, background levels in $[0.05,0.15]$, and smoothing scales in $[1,2]$ pixels.
\ptype{Low Confidence} uses the respective ranges $[0.4,0.6]$, $[0,0.05]$, and $[0.25,0.5]$.
\ptype{Dilated Regions} uses a disk radius sampled from one to five pixels, retains the confidence levels of \ptype{Balanced Response}, and uses smoothing scales in $[0.5,1]$.
\ptype{Small-Component Omission} removes components below an area cutoff sampled from 80--320 pixels, then uses foreground levels in $[0.8,1]$, background levels in $[0,0.1]$, and smoothing scales in $[0.5,1]$.
A shared noise field combines independent Gaussian pixel noise with a Gaussian-smoothed noise field, with standard deviations sampled from $[0.01,0.03]$ and $[0.01,0.02]$, respectively.
The smooth field uses a scale in $[4,8]$ pixels and is standardized before rescaling, and we add the noise field to each prediction and clip to $[0,1]$.
On an empty mask, every prediction type therefore reduces to its background level plus the shared noise field.
These values stay below $0.5$, so all 60 empty-group pairs are both empty under the baseline cutoff.

\parhead{Cutoff-Equality Cases.}
After this synthesis, the predictions of the additional masks in the four nonempty groups receive cutoff-equality cases.
These cases place prediction values exactly at $0.5$, so that \trow{Fixed-Threshold Binary ($P\geq0.5$)} of \cref{tab:controlled_cases} can differ from \trow{Baseline}.
For each constructed prediction mask, we select its largest connected component and form a band one pixel inside and outside its boundary.
A $90$-degree arc of this band, centered at a phase sampled from $[-\pi,\pi)$ and shared across the four prediction types, receives confidence $0.5$ after smoothing, noise, and clipping.

\subsubsection{Protocol Variant Details}
In \cref{tab:controlled_cases}, \trow{Soft Grayscale Map} computes \gls{IoU} from the soft confusion masses in \cref{eq:soft_tp,eq:soft_fp,eq:soft_fn,eq:soft_tn} and assigns zero when the denominator is zero.
Rows \tname{Fixed-Threshold Binary ($t=0.25$)} and \tname{Fixed-Threshold Binary ($t=0.75$)} keep the strict inequality of \trow{Baseline}.
Rows \tname{Threshold-wise Mean}, \tname{Threshold-wise Maximum}, and \tname{Threshold-wise AUC} use thresholds $t=j/100$ for $j=0,\ldots,100$ with inclusive comparisons $P\geq t$.
\trow{Threshold-wise AUC} uses trapezoidal endpoint weights, whereas \trow{Threshold-wise Mean} weights all 101 thresholds equally.
In \trow{Micro \gls{IoU} ($t=0.5$), Aggregation After S3}, empty-group images add no intersection and only their false-positive pixels to the summed union.
Because the empty-group predictions are empty at $P>0.5$, these images leave the micro score unchanged and require no both-empty convention.

\subsubsection{Analysis of Threshold-Dependent Reporting}
The rows in \tgroup{Metric Reporting} of \cref{tab:controlled_cases} apply one score formula to the same predictions and differ only in the report, so a metric name alone is often insufficient.
For example, \gls{Fbeta} may refer to an adaptive-threshold scalar, a maximum over a threshold sweep, a mean over the sweep, or a raw curve.
The same ambiguity applies to every score defined on binarized predictions, including \gls{Em} and \gls{MSIoU}, because the binarization policy and the report are chosen independently of the score formula.
These reports correspond to different questions, including a single operating point, the best-case operating point, threshold robustness, and the complete operating behavior.
The five-stage mapping separates these cases without requiring a new formula for each reporting convention.

\begin{table}[t!]
  \centering
  \caption{Effects of per-image min-max normalization on the shared benchmark.
  Each entry averages the per-image \gls{IoU} of one prediction type over all 300 \gls{GT} masks, with both-empty pairs scored as $0$.}
  \label{tab:prediction_preprocessing}
  \resizebox{\linewidth}{!}{\begin{tabular}{@{}lccccc@{}}
\toprule
Protocol & {Balanced Response} & {Low Confidence} & {Dilated Regions} & {Small-Component Omission} & {Pairwise Reversals \vs Baseline} \\
\midrule
\multicolumn{6}{@{}l@{}}{\textit{Fixed-Threshold Binary ($t=0.5$)}} \\
Raw Map (Baseline) & 0.700 & 0.380 & 0.475 & 0.552 & 0 \\
Min-Max Normalized Map & 0.730 & 0.792 & 0.470 & 0.555 & 3 \\
\midrule
\multicolumn{6}{@{}l@{}}{\textit{Soft Grayscale Map}} \\
Raw Map & 0.185 & 0.201 & 0.182 & 0.254 & 3 \\
Min-Max Normalized Map & 0.198 & 0.280 & 0.194 & 0.262 & 4 \\
\bottomrule
\end{tabular}
}
\end{table}

\subsection{Analysis of Min-Max Normalization}
\label{sup:prediction_preprocessing}

Some evaluation toolkits apply per-image min-max normalization for predictions before S1, which rescales the confidence values that every S1 representation operates on.
This optional preprocessing is specified separately from metric design and is not included as a metric mechanism in the proposed framework.
To further examine how this preprocessing affects the protocol comparisons in \cref{subsec:protocol_sensitivity}, we rescale each nonconstant prediction so that its minimum and maximum become zero and one and leave constant predictions unchanged.
The rescaled map $P'$ is then binarized using $P'>0.5$ before binary \gls{IoU} scoring or scored with soft \gls{IoU} (\cref{tab:prediction_preprocessing}).
The rescaling preserves the ordering of pixel values but stretches the foreground confidence of \ptype{Low Confidence} toward one.
The foreground pixels of \ptype{Low Confidence} below $0.5$ therefore pass the cutoff after rescaling.
\ptype{Low Confidence} consequently moves from last to first under thresholding and from second to first under soft scoring.

\subsection{S2--S4 Protocol Settings}
\label{sup:stage_protocols}

\parhead{Target Extraction \& Matching Details.}
In \cref{tab:controlled_cases}, \trow{Separated Component + Region} applies the size-invariant extraction rule~\cite{SizeInvarianceVariants} with a minimum component area of 50 pixels.
Each box is scored against all \gls{GT} foreground inside it, so it can include foreground from other \gls{GT} components, whether they are retained or fall below the area cutoff.
\trow{Whole Map + Edge} extracts inner boundary pixels by subtracting four-connected erosion from each binary mask, treating pixels outside the image as background.
Rows \tname{Threshold-only Matching ($\operatorname{IoU}>0.5$)}, \tname{Globally Optimized Matching}, and \tname{Nearest-neighbor Matching} use four-connected prediction and GT components.
For \trow{Threshold-only Matching ($\operatorname{IoU}>0.5$)}, an \gls{IoU} strictly above $\frac{1}{2}$ uniquely determines accepted one-to-one pairs when prediction components are mutually non-overlapping and GT components are mutually non-overlapping.
\trow{Globally Optimized Matching} applies Hungarian assignment to maximize the total component \gls{IoU} and keeps only assigned pairs with positive overlap.
\trow{Nearest-neighbor Matching} independently assigns every prediction component and every \gls{GT} component to its spatially nearest counterpart.
This row measures the distance between component centroids, which is one of several possible distances for region components, and resolves equal distances by the smaller component label.

\parhead{Score Computation Details.}
Rows \tname{Contribution Weighting}, \tname{Contextual Aggregation}, and \tname{Contribution Weighting + Contextual Aggregation} apply the contextual error $\widetilde E$ and the weights $W$ in \cref{eq:weighted_f_weights} of \gls{wFbeta} to the binary error map $E=|B-G|$, where contextual aggregation uses the normalized $7\times7$ Gaussian kernel of standard deviation 5 pixels with zero padding.
With either operation disabled, $\widetilde E=E$ or $W=1$, and the effective error is $E_e=W\widetilde E$.
The effective counts $\mathrm{FN}_e=\sum_{x:G(x)=1}E_e(x)$, $\mathrm{FP}_e=\sum_{x:G(x)=0}E_e(x)$, and $\mathrm{TP}_e=|G|-\mathrm{FN}_e$ give the score $\mathrm{TP}_e/(\mathrm{TP}_e+\mathrm{FP}_e+\mathrm{FN}_e)$.
Empty GT falls back to ordinary whole-map \gls{IoU}, and both-empty pairs score zero in every S2--S4 row, including \trow{Structural Statistics (\gls{Sm})}.

\parhead{Analysis of the Remaining S2--S4 Rows.}
This paragraph analyzes the rows of \cref{tab:controlled_cases} that \cref{subsec:protocol_sensitivity} leaves undiscussed.
In \trow{Whole Map + Skeleton}, skeleton overlap narrows the gaps among the three prediction types that retain most targets and places \ptype{Small-Component Omission} first and \ptype{Balanced Response} third.
The S3 and S4 choices change score magnitudes without reordering the prediction types on this benchmark.
The S3 rows match connected components, so they also use the separated-component granularity of S2-A2.
Globally optimized and nearest-neighbor matching lower \ptype{Low Confidence} most, because thresholding its weak foreground fragments targets into many small components.
Threshold-only matching further rejects weak component overlaps, which mainly lowers \ptype{Dilated Regions} and leaves \ptype{Small-Component Omission} unchanged.
Structural statistics in \gls{Sm} raise all scores while giving the baseline ranking.
We also transfer the contribution weights and contextual error aggregation of \gls{wFbeta}~\cite{wFmeasure} to the \gls{IoU} score through the effective counts defined above.
Contribution weighting lowers \ptype{Dilated Regions} by penalizing background errors farther from the \gls{GT}, and contextual aggregation raises \ptype{Low Confidence} by reducing effective errors near correctly predicted target pixels.
In \tgroup{Empty-Case Handling}, scoring both-empty pairs as one or excluding them shifts all scores.
Every prediction type behaves identically on the empty images, so the ranking cannot change in this benchmark, and absolute score comparisons still require the same empty-case convention.

\subsection{Matching Controls}
\label{sup:matching_constructions}

\parhead{Construction Details.}
The matching controls summarized in \cref{subsubsec:mechanism_attribution} use two equal squares of side 18--30 pixels and a six-column gap (\cref{tab:matching_experiments}).
Rows \tname{Bridge (1 px)}, \tname{Bridge (2 px)}, and \tname{Bridge (4 px)} fill the gap with a horizontal band of the listed width, and Rows \tname{Split (1 px)}, \tname{Split (2 px)}, and \tname{Split (4 px)} remove a vertical band of the same width through one square.
\trow{Missing Component} removes one square, and \trow{Extra Component} adds a disjoint $5\times5$ square.

\parhead{Analysis of Target Granularity \& Matching (S2-A, S3).}
Keeping region targets, we replace the whole-map target with connected components and compare the four matching mechanisms on merged, split, missing, and extra targets (\cref{tab:matching_experiments}).
The component-level mechanisms average assigned-pair \gls{IoU} over all prediction and \gls{GT} components, with zero contributions from unmatched components.
This shared score accommodates both one-to-one and many-to-one correspondence.
Bridging two targets leaves pixelwise \gls{IoU} high, while threshold-only matching rejects the resulting weak component overlaps.
Globally optimized matching accepts a partial overlap but can assign the merged prediction to only one \gls{GT} component.
Nearest-neighbor matching gives a higher component score because both \gls{GT} components can select the same prediction.
Splitting a target likewise gives a higher score under nearest-neighbor matching than under one-to-one assignment because several fragments can select the same \gls{GT} component.
Missing targets and detached extra components lower all three component-level scores through absent or zero-overlap correspondences.
\tgroup{Threshold-Induced Bridge} traces an S1 choice into S2 and S3 on one soft map whose two targets are joined by a lower-confidence bridge.
A low cutoff keeps the bridge and reproduces the merged configuration, a midpoint cutoff removes the bridge and restores both matches, and a cutoff at the target confidence removes both targets.
The same extraction, assignment, and scoring rules thus receive different entities after a threshold change in S1, while whole-map \gls{IoU} barely registers the merge.

\subsection{Error-Type Controls}
\label{sup:error_type_constructions}

\begin{table}[t!]
  \centering
  \caption{Variation in metric responses across sampled \gls{GT} masks.
  Each entry gives the mean and sample standard deviation over 300 sampled GT masks, with distances in pixels.}
  \label{tab:failure_variability}
  \resizebox{\linewidth}{!}{\begin{tabular}{@{}lccccccccccc@{}}
\toprule
Error Type & \gls{IoU} $\uparrow$ & \gls{MAE} $\downarrow$ & \glsxtrshort{wFbeta} $\uparrow$ & \glsxtrshort{Sm} $\uparrow$ & \glsxtrshort{Em} $\uparrow$ & \glsxtrshort{MSIoU}$_{J=1}$ $\uparrow$ & \glsxtrshort{MSIoU}$_{J=10}$ $\uparrow$ & \gls{ASD} $\downarrow$ & \gls{ASSD} $\downarrow$ & \gls{HD} $\downarrow$ & \gls{HD95} $\downarrow$ \\
\midrule
Thin Shift (1 px) & $0.500$\tsd{0.000} & $0.009$\tsd{0.000} & $0.740$\tsd{0.000} & $0.748$\tsd{0.000} & $0.993$\tsd{0.000} & $0.514$\tsd{0.000} & $0.933$\tsd{0.056} & $0.983$\tsd{0.000} & $0.983$\tsd{0.000} & $1.000$\tsd{0.000} & $1.000$\tsd{0.000} \\
Thin Shift (3 px) & $0.000$\tsd{0.000} & $0.027$\tsd{0.000} & $0.000$\tsd{0.000} & $0.464$\tsd{0.000} & $0.979$\tsd{0.000} & $0.497$\tsd{0.000} & $0.850$\tsd{0.076} & $2.000$\tsd{0.000} & $2.000$\tsd{0.000} & $3.000$\tsd{0.000} & $3.000$\tsd{0.000} \\
Thin Shift (6 px) & $0.000$\tsd{0.000} & $0.027$\tsd{0.000} & $0.000$\tsd{0.000} & $0.464$\tsd{0.000} & $0.979$\tsd{0.000} & $0.002$\tsd{0.000} & $0.663$\tsd{0.105} & $5.000$\tsd{0.000} & $5.000$\tsd{0.000} & $6.000$\tsd{0.000} & $6.000$\tsd{0.000} \\
Interior Removal (50\% of $G$) & $0.500$\tsd{0.000} & $0.049$\tsd{0.013} & $0.705$\tsd{0.004} & $0.741$\tsd{0.007} & $0.789$\tsd{0.006} & $1.000$\tsd{0.000} & $1.000$\tsd{0.000} & $2.780$\tsd{0.359} & $1.763$\tsd{0.226} & $7.087$\tsd{0.922} & $6.984$\tsd{0.900} \\
Missing $8\times8$ Component of $G$ & $0.973$\tsd{0.007} & $0.002$\tsd{0.000} & $0.987$\tsd{0.003} & $0.988$\tsd{0.003} & $0.998$\tsd{0.000} & $0.867$\tsd{0.013} & $0.832$\tsd{0.014} & $0.000$\tsd{0.000} & $5.009$\tsd{1.585} & $70.074$\tsd{11.044} & $66.442$\tsd{11.386} \\
Detached FP Pixel & $1.000$\tsd{0.000} & $0.000$\tsd{0.000} & $1.000$\tsd{0.000} & $0.998$\tsd{0.000} & $1.000$\tsd{0.000} & $1.000$\tsd{0.000} & $1.000$\tsd{0.000} & $0.382$\tsd{0.127} & $0.192$\tsd{0.064} & $64.460$\tsd{11.034} & $0.000$\tsd{0.000} \\
Detached $5\times5$ FP Patch & $0.989$\tsd{0.003} & $0.001$\tsd{0.000} & $0.989$\tsd{0.003} & $0.990$\tsd{0.001} & $0.999$\tsd{0.000} & $1.000$\tsd{0.000} & $1.000$\tsd{0.000} & $5.598$\tsd{1.771} & $2.931$\tsd{0.949} & $67.266$\tsd{11.040} & $63.699$\tsd{11.446} \\
Four Spurious Disks ($r=9$ px) & $0.706$\tsd{0.053} & $0.039$\tsd{0.000} & $0.706$\tsd{0.053} & $0.838$\tsd{0.027} & $0.915$\tsd{0.021} & $1.000$\tsd{0.000} & $1.000$\tsd{0.000} & $30.550$\tsd{6.109} & $20.852$\tsd{4.860} & $76.504$\tsd{7.816} & $72.159$\tsd{7.798} \\
\bottomrule
\end{tabular}
}
\end{table}

\parhead{Construction Details.}
For the error-type controls summarized in \cref{subsubsec:mechanism_attribution}, region masks for the error types in \cref{fig:controlled_perturbations} alternate rectangles and ellipses, centered in $[65,94]^2$ with half-extents or radii 20--33 pixels.
Rows \tname{Thin Shift (1 px)}, \tname{Thin Shift (3 px)}, and \tname{Thin Shift (6 px)} use vertical $116\times3$ rectangles shifted horizontally by the listed distance with zero padding, and the two larger shifts have zero overlap and identical TP, FP, FN, and TN for each sampled GT mask.
\trow{Interior Removal (50\% of $G$)} deletes half of the GT pixels in descending Euclidean distance from the background, with row-major ordering to break ties.
\trow{Missing $8\times8$ Component of $G$} adds an $8\times8$ GT square starting at $(8,8)$.
Rows \tname{Detached FP Pixel}, \tname{Detached $5\times5$ FP Patch}, and \tname{Four Spurious Disks ($r=9$ px)} add the pixel $(12,12)$, a predicted square starting at $(10,10)$, or disks centered at $(15,15)$, $(15,140)$, $(140,15)$, and $(140,140)$, respectively.
Columns \tname{\glsxtrshort{MSIoU}$_{J=10}$} and \tname{\glsxtrshort{MSIoU}$_{J=1}$} use the default $J=10$ levels ($j=0,\ldots,J-1$) and only the finest level ($j=0$), respectively.
\cref{tab:failure_experiments} reports means, and \cref{tab:failure_variability} reports sample standard deviations over the 300 sampled GT masks.

\parhead{Analysis of Target Type (S2-B).}
Each binary prediction copies a sampled \gls{GT} mask and introduces one of the error types in \cref{fig:controlled_perturbations}, and all metrics score the same predictions.
Keeping whole-map granularity and pixelwise matching, we compare region targets with the edge target of \gls{MSIoU} in \cref{tab:failure_experiments}.
Removing half of a target interior lowers the region scores but leaves \glsxtrshort{MSIoU}$_{J=1}$ and \glsxtrshort{MSIoU}$_{J=10}$~\cite{MSIoU} perfect, because their whole-map edge target retains only the outer boundary.
Adding detached false positives likewise reduces \gls{IoU} but leaves \gls{MSIoU} unchanged, because its \gls{GT}-edge recall does not penalize predicted edges outside the \gls{GT}.
A missing \gls{GT} component, in contrast, lowers \gls{MSIoU} because its boundary is no longer covered.

\parhead{Analysis of Thin Shifts (S4-A1--A2, S4-B1--B3).}
Confusion-statistics scoring (S4-A1) explains the responses of \gls{IoU} and \gls{MAE} to the thin shifts in \cref{tab:failure_experiments}, where region \gls{IoU} reaches zero once a thin target is displaced by its width and whole-image \gls{MAE} stays small because the error occupies little of the image.
\gls{Sm} and binary \gls{Em} also stay constant for shifts beyond the thin-target width, so they cannot distinguish further displacement.
\gls{wFbeta} still reaches zero once a thin target is displaced by its width, because both modifiers only rescale existing errors and cannot create true positives.
Pyramid processing (S4-B1) lets \gls{MSIoU} decrease gradually with the shift because coarser levels bring the displaced edges into common cells, whereas \glsxtrshort{MSIoU}$_{J=1}$ collapses once the displaced edges no longer coincide.

\parhead{Surface-Distance Details.}
Surface-distance boundary points are extracted as in \trow{Whole Map + Edge} of \cref{tab:controlled_cases}, and both boundaries are nonempty in every construction.
\cref{tab:surface_protocols} compares \gls{ASSD} and \gls{HD95} in \cref{eq:assd,eq:hd95} with the alternative summaries in \cref{sup:surface_distance}.
The 16 boundary points added in \trow{Detached $5\times5$ FP Patch} exceed five percent of the predicted boundary for every sampled GT mask, but fall below five percent of all boundary points for all 150 rectangular targets and 58 of the 150 elliptical targets, whose boundaries are long enough.

\parhead{Analysis of Spatial-distance Statistics (S4-A5).}
In \cref{tab:failure_experiments}, distances between matched boundary points grow with the thin shift, and \gls{HD} equals the shift, attained at the outer edges of the two thin rectangles (\cref{fig:surface_distance_controls}).
The distance summaries share binary masks, boundary pixel centers, nearest-neighbor correspondences, and unit pixel spacing, so their remaining differences isolate the summary rule.
Omitting a distant component, as in \trow{Missing $8\times8$ Component of $G$}, leaves directed \gls{ASD} at zero because every predicted boundary point still lies on the \gls{GT} boundary (\cref{fig:surface_distance_controls}).
\gls{ASSD} and \gls{HD} increase through the distances from \gls{GT} to prediction.
An isolated false-positive pixel produces a large \gls{HD} but leaves \gls{HD95} at zero because too few boundary points are affected to reach the percentile.
A $5\times5$ patch affects enough predicted boundary points to enter the directional percentile (\cref{fig:surface_distance_controls}).
Taking the percentile over all boundary points, as in \cref{tab:surface_protocols}, dilutes the affected fraction, so the patch no longer enters the percentile for 208 of the 300 sampled \gls{GT} masks, mostly those with longer target boundaries (\cref{fig:surface_distance_controls}).
Percentile scoring therefore reduces sensitivity to isolated extreme errors but can conceal a small erroneous component, and computing it over all boundary points makes this concealment depend on target size.

\begin{table}[t!]
  \centering
  \caption{Effects of directional aggregation on summaries of the same boundary distances.
  Each entry is the mean over 300 sampled GT masks, with distances in pixels.}
  \label{tab:surface_protocols}
  \resizebox{0.8\linewidth}{!}{\begin{tabular}{@{}lccc@{}}
\toprule
Summary & Missing $8\times8$ Component of $G$ & Detached FP Pixel & Detached $5\times5$ FP Patch \\
\midrule
\gls{ASD} $B\to G$ & 0.000 & 0.382 & 5.598 \\
\gls{ASD} $G\to B$ & 9.283 & 0.000 & 0.000 \\
\midrule
Mean Over All Boundary Points (\gls{ASSD}) & 5.009 & 0.192 & 2.931 \\
Mean of Directional Means & 4.641 & 0.191 & 2.799 \\
\midrule
Max of Directional Percentiles (\gls{HD95}) & 66.442 & 0.000 & 63.699 \\
Percentile Over All Boundary Points & 63.126 & 0.000 & 15.922 \\
\bottomrule
\end{tabular}
}
\end{table}

\subsection{Scoring Controls}
\label{sup:scoring_controls}

\parhead{Fixed-Count Layout Details.}
The fixed-count controls in \cref{fig:scoring_controls} place a $96\times96$ \gls{GT} square with its upper-left corner at $(32+d_y,58+d_x)$, where $d_y\in[-4,4]$ and $d_x\in[-2,2]$.
The merged TP area is a $44\times64$ rectangle rotated by $30^\circ$ about the target center and rasterized at pixel centers.
The split TP area cuts this rectangle along its two midlines and translates each piece by integer offsets toward the nearest target corner.
Both TP areas stay at least six pixels from the target boundary, and every remaining target pixel is FN.
A $32\times32$ FP block occupies rows 32--63 relative to the target and either ends at its left boundary or is separated from it by 20 columns.
Every layout contains 2,814 TP, 1,024 FP, 6,402 FN, and 15,360 TN pixels.
\gls{HCE} uses five relaxation steps and a polygon simplification tolerance of two pixels, and the FN region and the FP block survive the relaxation in every layout.

\begin{table}[t!]
  \centering
  \caption{Effects of RGB background on context-based scoring and camouflage-derived contribution weighting.
  Each entry is the mean over 300 sampled GT masks with $\beta^2=1$, and only the RGB background differs across rows.}
  \label{tab:context_controls}
  \resizebox{0.8\linewidth}{!}{\begin{tabular}{@{}lccccc@{}}
\toprule
Background & {Forward Response} & {Reverse Response} & {Weighted Reverse Response} & \gls{Cm} & \gls{wCm} \\
\midrule
Neutral & 0.471 & 0.498 & 0.500 & 0.484 & 0.485 \\
Covered-Side Match & 0.471 & 0.498 & 0.579 & 0.484 & 0.519 \\
Missed-Side Match & 0.471 & 0.498 & 0.429 & 0.484 & 0.449 \\
\bottomrule
\end{tabular}
}
\end{table}

\parhead{Analysis of RGB Background Controls.}
\label{sup:context_controls}
To isolate camouflage-derived contribution weighting, we vary only the RGB background for \gls{Cm} and its weighted variant \gls{wCm}, fixing the soft prediction, \gls{GT}, and foreground colors (\cref{tab:context_controls}).
The controls use a rectangular target of independently sampled height and width 32--40 pixels, with its upper-left corner at $(62,62)$ plus coordinate offsets in $[-4,4]$.
Predictions are $0.85$ on the left half, $0.15$ on the right half, and $0.02$ outside GT.
The covered left half and the missed right half have fixed RGB colors $(50,150,50)$ and $(180,50,180)$.
Rows \tname{Covered-Side Match}, \tname{Missed-Side Match}, and \tname{Neutral} set the background to $(50,150,50)$, $(180,50,180)$, and $(110,110,110)$, respectively.
Both \gls{Cm} and \gls{wCm} use $\beta^2=1$, which replaces the default of $1.2$ for \gls{wCm}.
The base \gls{Cm} remains unchanged because its forward and reverse responses (S4-A4) depend only on the prediction and \gls{GT}.
Matching the background color to the better-covered target half raises the weighted reverse response and \gls{wCm}, while matching the poorly covered half lowers both.
The RGB change therefore acts only through the camouflage-derived weights in \cref{eq:context_weighted_reverse}.

\subsection{Composed-Metric Tests}
\label{sup:composition_constructions}

\parhead{Test Layout Details.}
Each of the 300 GT layouts in \cref{subsubsec:composed_metric_evaluation} contains two separated rectangles with a shared sampled height of 12--18 pixels, and the smaller width is sampled from $\{6,8\}$ pixels.
The larger rectangle is five times as wide.
The upper-left corners of the larger and smaller rectangles are $(20+d_y,20+d_x)$ and $(70+d_y,120+d_x)$, respectively, where $d_y$ and $d_x$ are sampled independently from $\{0,\ldots,11\}$.

\parhead{Paired-Test Prediction Details.}
For the paired tests in \cref{tab:composition_experiments}, the pair in \tgroup{Confidence} predicts the \gls{GT} with confidence $0.75$ or $0.25$, on opposite sides of the fixed cutoff.
The pair in \tgroup{Target Coverage} misses the same number of pixels, either as a strip from each target or as the whole smaller target, and each strip has half the smaller width.
The pair in \tgroup{Background FP} predicts both targets exactly and adds remote false positives covering one quarter or all of the \gls{GT} area.
These false positives fill nested subsets of a remote area, rows 120--154 and columns 5--154 in zero-based indices, in raster order.
The pair in \tgroup{Displacement} thins both targets to three pixels and shifts the predictions by 3 or 24 pixels, so neither prediction overlaps \gls{GT}.
The pair in \tgroup{Combined Errors} either shifts both targets by three pixels at confidence $0.7$, or omits the smaller target and adds remote false positives of the same area.
In \tgroup{Combined Errors}, the prediction with an omitted target assigns confidence one to the larger rectangle.
All unspecified pixels match GT, and no shift clips a prediction at the image boundary.
The preference in \tgroup{Combined Errors} expresses the chosen importance of retaining both targets with a small spatial error over omitting a target and introducing remote FP.
\trow{Pairs Meeting the Requirement (of 300)} compares the two predictions within each pair before averaging, so it cannot be read from the mean scores that the other rows of \cref{tab:composition_experiments} report for each prediction separately.

\parhead{Analysis of Grid-Origin Sensitivity.}
For the regional pyramids of \gls{ETMSIoU}, we pad zeros at the top and left to make the map dimensions divisible by each cell side $2^j$, $j=0,\ldots,J-1$, following the \gls{MSIoU} grid convention~\cite{MSIoU}.
The top-left zero padding aligns every pooling grid with the bottom-right image corner, so appending zero rows and columns at the bottom and right moves the grid relative to the unchanged maps.
In the 23 pairs of \tgroup{Combined Errors} that do not meet the requirement in \cref{tab:composition_experiments}, coarse cells pool the remote FP with the omitted target and grant it agreement, and the grid alignment lets this credit outweigh the penalty of the small shift.
Over 16 grid origins shifted by 0, 32, 64, or 96 pixels in each direction, \tgroup{Combined Errors} yields between 0 and 56 exceptions, compared with the 23 exceptions at the default origin in \cref{subsubsec:composed_metric_evaluation}, and no pair fails at every origin.
The four single-requirement tests hold at every origin.

\newpage
\section{Extended Discussion}
\label{sup:extended_discussion}

This section extends \cref{sec:discussion} with metric-level interpretation, the practical cost of unstated protocols, and new mechanisms within the five stages.

\subsection{Metric Interpretation and Comparison}

\parhead{Metrics as Protocols.}
The five-stage analysis treats a metric as a complete evaluation protocol.
The same final score can behave differently depending on whether the prediction is kept soft or binarized, whether the target is evaluated as a whole map or decomposed into entities, whether correspondence is implicit or explicitly matched, and whether the result is reported as a scalar, a curve, or an area summary (\cref{subsec:protocol_sensitivity}).
The five-stage framework records where these choices enter the protocol, and each metric instantiates only the mechanisms it needs.

\parhead{Question-Dependent Comparison.}
Metric comparison should be tied to the question being asked.
\gls{MAE} is appropriate when soft pixel disagreement is the main concern, while \gls{Fbeta}, \gls{IoU}, Dice, and \gls{BER} summarize different views of thresholded confusion statistics.
\gls{wFbeta} modifies the underlying error evidence and its contributions, while \gls{Sm} uses object and spatial block statistics.
\gls{Em} expresses local-global alignment as a nonlinear function of the four confusion counts.
\gls{MSIoU}, \gls{hIoU}, \gls{HCE}, and \gls{Cm} move further by changing the evaluated entity, the matching relation, the correction cost, or the contextual response, and \gls{wCm} adds image-conditioned difficulty.
These differences correspond to different claims about what should count as good segmentation.

\parhead{Reporting Rules.}
Thresholding and metric reporting remain central sources of ambiguity.
A fixed threshold approximates a deployment decision, an adaptive threshold follows the confidence scale of each prediction map, and a dynamic threshold sweep evaluates a sequence of operating points.
Reporting a maximum, a mean, raw curve data, or an \gls{AUC} changes the scientific question from best-case operation to threshold robustness, curve inspection, or integrated behavior.
This distinction is especially important for metrics built from continuous maps, where ranking behavior and absolute confidence scale are not the same thing.
A strong benchmark should state the threshold protocol and the reporting rule as part of the metric.

\parhead{Target Representation.}
Target representation is another point where a metric silently encodes a task assumption.
Whole-map metrics are simple and stable, but they mainly evaluate region agreement after spatial alignment.
Size-invariant variants reduce sample-level aggregation bias by giving connected targets more balanced influence.
\gls{hIoU} makes target localization explicit for \gls{IRSTD}, where extremely small targets, sparse distributions, and strong distractors make object-level localization a visible part of performance.
The open issue is whether the evaluated entities match the operational meaning of the task.

\parhead{Beyond Overlap.}
Structural, boundary, contextual, and effort-aware metrics show that binary target segmentation quality extends beyond overlap.
\gls{Sm} evaluates object and region structure, \gls{MSIoU} emphasizes edge coverage across pyramid levels, and \gls{HCE} estimates prescribed correction operations.
\gls{Cm} combines contextual directional responses, and its weighted extension introduces image-derived camouflage difficulty.
These metrics are useful because they make different failure modes visible.
Each added mechanism also introduces assumptions about structure, scale, editing cost, or visual context.
Such assumptions should be stated explicitly, especially when a metric is transferred from one task family to another.

\parhead{Metric Combinations.}
The controlled analyses in \cref{sup:error_type_constructions} show that region overlap and boundary distance capture complementary aspects of the same errors.
The former detects missing interiors and false-positive area, while \gls{ASSD} and \gls{MSIoU} still distinguish displacements after overlap is lost.
\gls{HD} and unmatched-component counts preserve sensitivity to isolated distant errors that \gls{HD95} may discard.
\gls{HCE} adds the number and type of correction operations, which count-based scores ignore by construction.
Metric combinations can therefore address displacement, region fidelity, and repair effort through the evidence retained at each stage.

\subsection{Benchmark Reporting and Reproducibility}

\parhead{Protocol-Dependent Scores.}
As shown in \cref{subsec:protocol_sensitivity}, threshold selection, the use of $>$ or $\geq$, target extraction, matching, empty-target handling, and the aggregation position can all change reported scores, and per-image min-max normalization has the same effect (\cref{sup:prediction_preprocessing}).

\parhead{Re-Evaluation Burden.}
When protocol details are missing, researchers often have to collect prediction maps and re-evaluate them under the same metric protocol.
Such repeated evaluation is costly, and it makes comparisons depend on which prediction maps are available, which datasets remain accessible, and which protocol version is chosen.
When a benchmark reports only a mean score and a compact formula, the scoring rule is usually visible, but the surrounding protocol choices can fade into the background.
Prediction representation, thresholding, target extraction, matching, and reporting conventions may then be left implicit even though they affect the reported result.
Shared evaluation protocols help, and each paper still needs to specify the protocol it uses.

\parhead{Dataset Aggregation.}
Benchmark tables should state the aggregation position defined in \cref{subsec:dataset_aggregation}, together with the included images and their weights.
Beyond \gls{IoU}, micro aggregation after S3 applies to any metric whose score quantities can be combined across images.

\subsection{New Mechanisms Within the Five Stages}

The following examples illustrate new mechanisms that each stage could host.
S1 could introduce confidence-scale-aware or uncertainty-aware prediction representations, S2 could define topology units, holes, branches, part regions, or uncertainty bands as target entities, and S3 could introduce tolerance-aware, part-aware, or temporal matching when the task requires it.
S4 could add scoring terms for topology preservation or application-specific functional error, while S5 could preserve richer reporting forms such as summaries for different groups, curves, or dataset-level distributions.
Even a substantially new metric can be described as changing one or more stages.
The design space is useful only when the changed stage is made explicit.

\newpage
\bibliography{string2full,paper}

\end{document}